\pdfoutput=1
\documentclass{article}

\usepackage[preprint]{neurips_2026}
\usepackage{amsmath}
\usepackage{mathtools}
\usepackage{booktabs}
\usepackage{multirow}
\usepackage{makecell}
\usepackage{tabularx}
\usepackage{array}
\usepackage{ragged2e}
\usepackage{caption}
\usepackage{graphicx}
\usepackage{float}
\usepackage{placeins}
\usepackage{bm}
\usepackage{tabularx}

\usepackage{threeparttable}
\usepackage{pifont}
\newcommand{\cmark}{\ding{51}}
\newcommand{\xmark}{\ding{55}}

\newcolumntype{L}[1]{>{\RaggedRight\arraybackslash}p{#1}}
\newcolumntype{Y}{>{\RaggedRight\arraybackslash}X}

\makeatletter
\renewcommand{\section}{%
  \@startsection{section}{1}{\z@}%
                {-1.4ex \@plus -0.4ex \@minus -0.2ex}%
                { 0.9ex \@plus  0.2ex \@minus  0.1ex}%
                {\large\bf\raggedright}%
}
\makeatother
\usepackage[utf8]{inputenc} 
\usepackage[T1]{fontenc}    
\usepackage{hyperref}       
\usepackage{url}            
\usepackage{booktabs}       
\usepackage{amsfonts}       
\usepackage{amssymb}
\usepackage{nicefrac}       
\usepackage{microtype}      
\usepackage{enumitem}
\usepackage[table]{xcolor}  
\renewcommand{\cmark}{\textcolor{teal!70!black}{\ensuremath{\checkmark}}}
\renewcommand{\xmark}{\textcolor{red!80!black}{\ensuremath{\times}}}

\newcommand{\benchcell}[2]{#1\hspace{0.35em}{\scriptsize\cite{#2}}}

\title{\texorpdfstring{%
\parbox{\textwidth}{%
\centering
\makebox[\textwidth][c]{%
\raisebox{-0.08\height}{\includegraphics[width=0.075\textwidth]{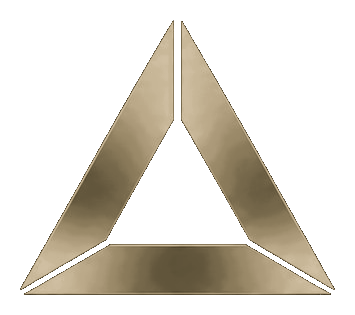}}\hspace{0.55em}%
{\fontsize{20}{24}\selectfont\bfseries SkillAudit}%
}\\[0.55em]
{\fontsize{15}{17}\selectfont From Fixed-Suite Benchmarking to Skill-Centered Assessment}%
}}{SkillAudit: From Fixed-Suite Benchmarking to Skill-Centered Assessment}}

\author{%
\normalfont
\makebox[\textwidth][c]{%
\parbox{1.0\textwidth}{%
\centering
\small
\textit{Dexu Yu}\textsuperscript{*,1},
\textit{Youhua Li}\textsuperscript{*,2,11},
\textit{Zhaoyang Guan}\textsuperscript{*,3},
\textit{Xianhao Lin}\textsuperscript{*,4,11},
\textit{Jining Luan}\textsuperscript{5,11},
\textit{Zihao Rao}\textsuperscript{5,11},
\textit{Xuanqi Lan}\textsuperscript{6},
\textit{Yang Ran}\textsuperscript{7},
\textit{Bo Lan}\textsuperscript{7},
\textit{Nai-Xin Zhai}\textsuperscript{2,11},
\textit{Hanwen Du}\textsuperscript{8,11},
\textit{Junchen Fu}\textsuperscript{9,11},
\textit{Wenhao Deng}\textsuperscript{9,11}, \\
\textit{Yongxin Ni}\textsuperscript{10,\ensuremath{\dagger},\S},
\textit{Chunxiao Li}\textsuperscript{5,\ensuremath{\dagger}}
\\
\scriptsize
\textsuperscript{1}Northeastern University,
\textsuperscript{2}City University of Hong Kong,
\textsuperscript{3}Northwestern University,
\textsuperscript{4}Fudan University, \\
\textsuperscript{5}University of Science and Technology of China,
\textsuperscript{6}Santa Clara University,
\textsuperscript{7}Fenz AI, \\
\textsuperscript{8}Ohio State University,
\textsuperscript{9}University of Glasgow,
\textsuperscript{10}National University of Singapore,
\textsuperscript{11}DeciLix Lab \\
\footnotesize
\textsuperscript{*}Equal contribution.
\quad
\textsuperscript{\ensuremath{\dagger}}Corresponding authors\\
\footnotesize
\textsuperscript{}{\{ youhuali2-c@my.cityu.edu.hk, chunxiao.li@ustc.edu.cn, niyongxin@u.nus.edu\}}
}%
}%
}

\begin{document}

\maketitle

\begingroup
\renewcommand{\thefootnote}{\fnsymbol{footnote}}
\footnotetext[4]{Project Leader}
\endgroup

\begin{center}
  \small Homepage: {\color{blue!70!black}\url{https://skillaudit.github.io/}}
\end{center}

\begin{abstract}
Agent skills have become a practical way to extend large language model agents, but the growing skill ecosystem still lacks a reliable way to judge whether a skill is worth deploying. Existing evaluation methods remain largely anchored to fixed task suites, assessing skills through performance on predefined tasks and environments. As skill marketplaces expand, this paradigm becomes inadequate: fixed suites can conflate a skill's marginal contribution with backbone strength and miss its value when tasks fall outside the skill's intended scope. We introduce \textbf{SkillAudit}, an end-to-end framework for \emph{skill-centered assessment} that takes an arbitrary agent skill as input and automatically generates a comprehensive, multi-dimensional evaluation report spanning utility, efficiency/cost, and safety. 
SkillAudit focuses on the skill artifact itself and constructs capability-aligned evaluation tasks directly from the skill package. The generated tasks are conducted in isolated sandbox environments to collect execution evidence, followed by automated checks with LLM-based judging to produce auditable results. To dissect the agent skills, we propose the \emph{baseline comparison principle} to measure utility and efficiency/cost, and introduce a \emph{two-stage detection paradigm} combining static semantic analysis with dynamic runtime verification to assess safety risks. 
After scanning top-ranked real-world skill packages spanning 23 occupational categories, we found that over 7\% of skills are at risky status.
\textit{We surface SkillAudit through a browser extension for user access at discovery time, with intermediate trajectories and all evaluation artifacts open-sourced at {\color{blue!70!black}\url{https://github.com/SkillAudit/skillaudit}}}.
\end{abstract}


\section{Introduction}
\label{sec:intro}

Agent skills are becoming a common way to extend large language model agents. Skills are shareable artifacts that an agent can load on demand to specialize in a family of tasks~\cite{anthropic_skills_2025}. Unlike models or single tool calls, skills package procedural knowledge, executable code, and environment assumptions into self-contained units that can be reused. Typically, they consist of a \texttt{SKILL.md} file along with scripts, templates, dependencies, and metadata. Since Anthropic released the agent skill format in late 2025, public ecosystems have expanded from dozens of artifacts to hundreds of thousands across GitHub, community marketplaces, and corporate contributions~\cite{li2026skillsbench}. 


Despite agent skills increasingly resembling installable, software-like artifacts, developers still lack a principled way to judge whether a given skill is worth enabling. Adoption decisions today rely predominantly on README claims, star counts, and download figures surfaced by GitHub repositories and skill hubs, rather than on direct empirical evaluation evidence. This creates a critical evaluation gap at adoption time: it remains unclear whether a particular skill will improve task outcomes, reduce operational effort (time or tokens), or introduce new safety risks into the workflows it claims to support.

Existing benchmarks evaluate agents, tools, or skill-augmented performance on fixed tasks, but they do not provide a general evaluation framework that takes an arbitrary skill artifact as input and produces a detailed, evidence-grounded skill report. AgentBench~\cite{liu2023agentbench}, $\tau$-bench~\cite{yao2024taubench}, and TheAgentCompany~\cite{xu2025agentcompany} evaluate agents on fixed task suites; ToolLLM~\cite{qin2023toolllm} and BFCL~\cite{bfcl2024} study tool invocation behavior; InjecAgent~\cite{zhan2024injecagent} emphasizes adversarial robustness; and SkillsBench~\cite{li2026skillsbench} evaluates skill artifacts against a curated task suite. Table~\ref{tab:related} summarizes the features of existing benchmarks. The shared limitation is that tasks are fixed in advance, meaning such benchmarks can only effectively evaluate skills whose intended operating range already aligns with the predefined suite, leaving most real-world skills without an appropriate evaluation pathway.


\begin{figure*}[t]
\centering
\includegraphics[width=0.80\textwidth]{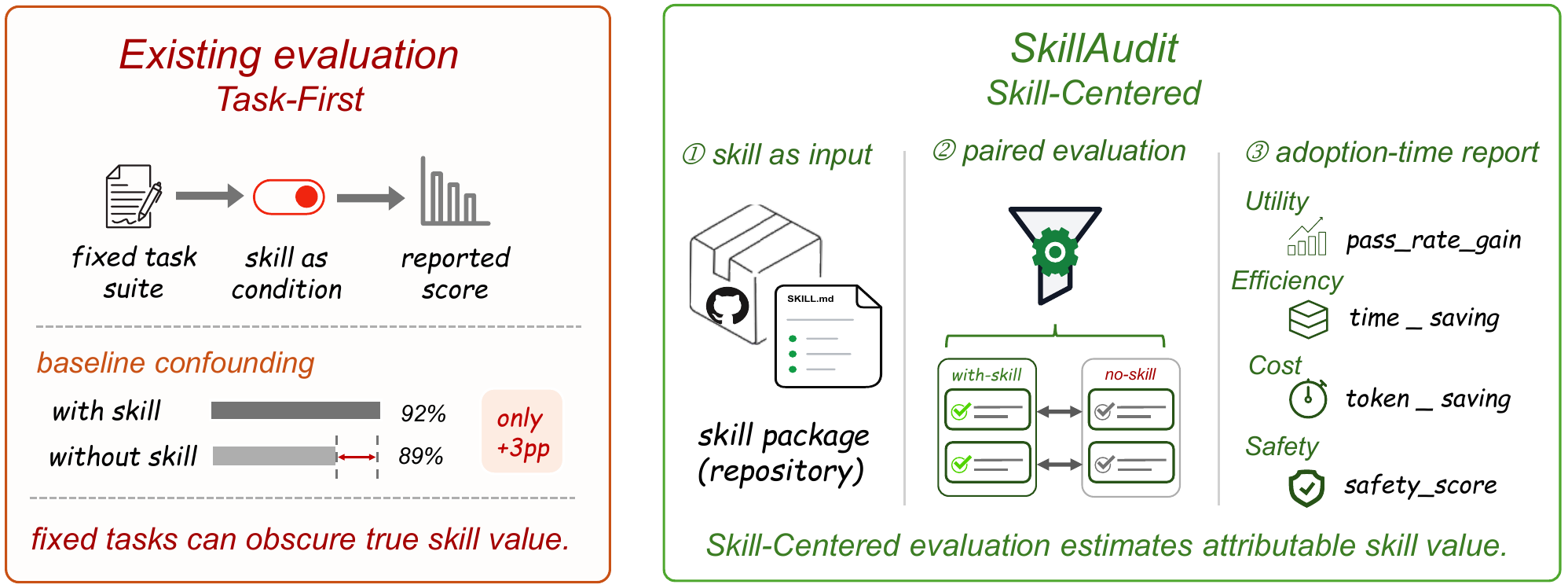}
\caption{SkillAudit centers evaluation on each skill artifact rather than benchmarking skills on a fixed task suite, providing a multi-dimensional evaluation report.}
\label{fig:motivation}
\end{figure*}

\begin{table*}[!t]
\centering
\small
\setlength{\tabcolsep}{4pt}
\renewcommand{\arraystretch}{1.02}
\caption{Comparison of SkillAudit with representative prior benchmarks across key evaluation properties. Check marks denote supported properties.}
\label{tab:related}
\begin{tabularx}{\textwidth}{@{}Y >{\centering\arraybackslash}m{1.10cm} >{\centering\arraybackslash}m{1.15cm} >{\centering\arraybackslash}m{1.00cm} >{\centering\arraybackslash}m{1.00cm} >{\centering\arraybackslash}m{1.15cm} >{\centering\arraybackslash}m{1.70cm}@{}}
\toprule
\textbf{Benchmark} & \shortstack[c]{\textbf{Skill}\\[-2pt]\textbf{artifact}} & \shortstack[c]{\textbf{Skill-}\\[-2pt]\textbf{centered}} & \textbf{Utility} & \textbf{Safety} & \textbf{Efficiency} & \shortstack[c]{\textbf{Adoption}\\[-2pt]\textbf{surfacing}} \\
\midrule
\benchcell{AgentBench}{liu2023agentbench}       & \xmark & \xmark & \cmark & \xmark & \xmark & \xmark \\
\benchcell{$\tau$-bench}{yao2024taubench}       & \xmark & \xmark & \cmark & \xmark & \xmark & \xmark \\
\benchcell{TheAgentCompany}{xu2025agentcompany} & \xmark & \xmark & \cmark & \xmark & \xmark & \xmark \\
\benchcell{ToolLLM}{qin2023toolllm}             & \xmark & \xmark & \cmark & \xmark & \xmark & \xmark \\
\benchcell{BFCL}{bfcl2024}                      & \xmark & \xmark & \cmark & \xmark & \xmark & \xmark \\
\benchcell{InjecAgent}{zhan2024injecagent}      & \xmark & \xmark & \xmark & \cmark & \xmark & \xmark \\
\benchcell{SkillsBench}{li2026skillsbench}      & \cmark & \xmark & \cmark & \xmark & \xmark & \xmark \\
\textbf{SkillAudit {\scriptsize(ours)}} & \textbf{\cmark} & \textbf{\cmark} & \textbf{\cmark} & \textbf{\cmark} & \textbf{\cmark} & \textbf{\cmark} \\
\bottomrule
\end{tabularx}
\end{table*}

We therefore introduce \textbf{SkillAudit}, a framework for \emph{skill-centered assessment} that takes an arbitrary skill as input and automatically generates a multi-dimensional evaluation report spanning \textbf{utility, efficiency/cost, and safety}. Rather than measuring how a skill shifts performance on a fixed benchmark, SkillAudit asks how a skill package can be systematically transformed into evaluation tasks that reflect what the skill is actually designed to do, as illustrated in Figure~\ref{fig:motivation}. To this end, SkillAudit first employs an LLM to parse \texttt{SKILL.md}, extracting the skill's core capabilities and operational workflow, and formulates an evaluation scheme precisely tailored to the target skill. To prevent the skill's procedural knowledge from being inadvertently encoded in the agent's task and thus invalidating the evaluation, the agent is exposed only to the business-goal instructions, while the evaluator retains the capability analysis, scoring rubric, and expected outcomes. SkillAudit then compiles each scheme into executable task instances, runs them in isolated sandboxed environments, and applies rule-based checks alongside LLM-as-judge scoring to produce auditable evaluation results. The comparison of the SkillAudit and some representative agent skill evaluation frameworks and tools is illustrated in Table~\ref{tab:comparison}.

\begin{table*}[t]
\centering
\caption{Comparison of representative agent skill evaluation frameworks.
\textbf{Evaluation Object}: the skill artifact, the pipeline that generates skills, or the skill reuse method.
\textbf{Skill Source}: a curated set or applicable to arbitrary skills in open ecosystem.
\textbf{Task Auto-gen.}: whether evaluation tasks are automatically generated from the skill itself rather than from a fixed suite.
\textbf{Sandbox}: whether tasks are executed in an isolated sandbox environment.
\textbf{Utility}: utility metric.
\textbf{Safety}: \emph{One layer} indicates static semantic analysis or dynamic runtime test only; \emph{Two layers} indicates both static and dynamic; $\times$ indicates none.
\textbf{Efficiency/Cost}: whether token or time savings are measured.
\textbf{Auditable Evidence}: whether verifiable execution evidence is provided for users.
\textbf{Adoption Surfacing}: whether evaluation results are surfaced at skill adoption or discovery time.
}
\label{tab:comparison}
\begin{threeparttable}
\resizebox{\textwidth}{!}{%
\begin{tabular}{p{3.2cm} p{2.0cm} p{2.0cm} p{1.2cm} p{1.2cm} p{2.8cm} p{1.6cm} p{1.2cm} p{1.2cm} p{1.4cm}}
\toprule
\textbf{System} &
\textbf{Evaluation \newline Object} &
\textbf{Skill Source} &
\textbf{Task \newline Auto-gen.} &
\textbf{Sandbox} &
\textbf{Utility} &
\textbf{Safety} &
\textbf{Efficiency \newline \& Cost} &
\textbf{Auditable \newline Evidence} &
\textbf{Adoption \newline Surfacing} \\
\midrule
SkillsBench~\cite{li2026skillsbench}
  & Skill artifact & Curated & \xmark & \cmark
  & Pass rate gain & \xmark & \xmark & \cmark & \xmark \\
SWE-Skills-Bench~\cite{han2026swe}
  & Skill artifact & Curated & \xmark & \cmark
  & Pass rate gain & \xmark & \cmark & \cmark & \xmark \\
SkillTester~\cite{wang2026skilltester}
  & Skill artifact & Arbitrary & \cmark & \xmark
  & Pass rate gain & One layer & \xmark & \cmark & Web UI \\
SkillGenBench~\cite{zhou2026skillgenbench}
  & Skill pipeline & Curated & \xmark & \xmark
  & Pass@$k$ & \xmark & \xmark & \xmark & \xmark \\
SkillLens~\cite{huang2026raw}
  & Skill pipeline & Curated & \xmark & \xmark
  & Extraction efficacy; \newline Target evolvability & \xmark & \xmark & \cmark & \xmark \\
SkillLens\tnote{$\dagger$}~\cite{miao2026skilllens}
  & Skill reuse & Curated & \xmark & \xmark
  & Acc@1; \newline Success rate & \xmark & \cmark & \xmark & \xmark \\
\midrule
\textbf{SkillAudit (ours)}
  & Skill artifact & Arbitrary & \cmark & \cmark
  & Pass rate gain & Two layers & \cmark & \cmark & Browser \newline extension \\
\bottomrule
\end{tabular}%
}
\begin{tablenotes}
\item[$\dagger$] {\scriptsize This SkillLens is not strictly an agent skill evaluation framework; included here to disambiguate from other works sharing the same name.}
\end{tablenotes}
\end{threeparttable}
\end{table*}

When assessing utility and efficiency/cost, focusing solely on the agent-with-skill setting conflates the agent's intrinsic capabilities with the skill's marginal contribution. To address this, we adopt the  \textbf{baseline comparison principle}. Specifically, each evaluation task is executed under two controlled conditions with identical configurations and scoring rubrics except for the presence of the target skill, i.e., paired with-skill and no-skill conditions. This design is motivated by the inherently relative nature of a skill's marginal contribution, which captures the incremental gain over the baseline agent. And as the baseline agent gets stronger, it may subsume what a skill offers, thereby decreasing the skill's marginal value. Regarding safety risks of a skill, we note that even if a safety risk is present in the skill's instructions or scripts, the agent does not necessarily trigger it due to its own security guardrails. The agent may recognize the risks and proactively refuse to execute the risky actions. Therefore, it is essential to distinguish between the \textit{existence} of a risk and its \textit{exploitability} in practice. To this end, we propose the \textbf{two-stage safety risk detection principle}. In the first stage, SkillAudit predefines 21 risk patterns and conducts a comprehensive static semantic analysis of the skill package to identify which patterns are present, corresponding to the risks' existence. In the second stage, SkillAudit constructs targeted runtime tests to determine whether the agent would actually trigger the identified risks during execution, assessing their exploitability. To validate the static semantic scanner in the first stage, we construct a controlled dataset of 29 skill packages with injected risk instances and human-annotated ground truth. On this dataset, the scanner achieves high recall on medium- and high-severity patterns ($96.0\%$ and $97.0\%$, respectively), demonstrating its reliability for large-scale deployment.

To connect assessment with real adoption behavior, SkillAudit also includes a browser extension for skill-hosting platforms. When a developer visits a skill page, the extension displays a SkillAudit action, submits or retrieves the corresponding assessment, and opens an interactive report. This turns skill assessment from an offline benchmarking exercise into an adoption-time decision aid.


Our contributions are fourfold.

\begin{itemize}[leftmargin=*, topsep=0pt]

    \item \textbf{An evidence-grounded evaluation framework for open skill ecosystems.} We introduce SkillAudit, an end-to-end framework that takes an arbitrary agent skill as input and automatically generates a comprehensive, multi-dimensional evaluation report. It derives capability-matched utility schemes and security probes, executes them in instrumented sandboxes, and grounds every reported judgment in concrete execution evidence spanning utility, efficiency/cost, and safety.

    \item \textbf{A skill-centered formulation with baseline comparison and two-stage risk detection.} We frame skill assessment around the skill artifact itself, adopting a baseline comparison principle that isolates marginal skill gain from backbone strength. For safety, we combine static semantic analysis with targeted dynamic runtime verification, jointly assessing both existence and exploitability of risks.

    \item \textbf{A discovery-time browser extension for practical skill adoption.} We integrate SkillAudit reports into skill-hosting pages, connecting benchmark evidence to the moment when developers actually decide whether to install a skill.

    \item \textbf{Empirical validation across real-world skills and agent configurations.} We evaluate SkillAudit on 226 real-world skill packages spanning 23 occupational categories across six agent--model configurations, producing a detailed evaluation report for each skill. Our key findings reveal that skill utility decreases as backbone strength increases, and that safety risk is largely independent of utility. Notably, over 7\% of evaluated skills are rated as risky. 
    
\end{itemize}

\section{Problem Formulation}
\label{sec:problem}

\subsection{Skill Artifact}
Agent skills are structured packages of procedural knowledge that augment agent capabilities at inference time without requiring model modification~\cite{li2026skillsbench}. Unlike prompts, RAG, or tool calls, a skill encapsulates domain knowledge, standard operating procedures, task-specific experience, and user preferences into an installable package centered on \texttt{SKILL.md} and accompanied by executable scripts and code templates.

The skill ecosystem is characterized by two coexisting properties: broad domain coverage and the high specialization of individual skills. As publicly available skills proliferate across diverse domains, each remains narrowly tailored to a specific task class with its own typical workflow and operational context. This makes it impractical to evaluate skills through a fixed set of predefined task categories or manually constructed tests. Automatically generating customized evaluation tasks on a per-skill basis thus emerges as the natural and necessary direction.

\subsection{The Adoption-Time Question}

As the public skill ecosystem grows to hundreds of thousands of packages, the question developers face is not which agent scores higher on a benchmark, but a practical adoption-time decision: \textit{given a skill package, should I install it?}

This question decomposes into three assessment dimensions.

\noindent\textbf{Utility.} What incremental gains does the skill provide over the agent's baseline competence? Utility reflects the extended capabilities offered by the skill. Notably, the marginal value a skill delivers varies across different agent configurations. 


\noindent\textbf{Efficiency and Cost.} Does the skill improve efficiency by reducing the time and steps required to complete a task, or lower cost by decreasing token consumption? We wonder whether the domain knowledge and experience of a skill can streamline the agent's execution path.

\noindent\textbf{Safety Risk.} Does the skill introduce potential security threats, such as prompt injection, data exfiltration, hidden instructions, or privilege escalation? Malicious skills can expose users to serious harm, making safety one of the most critical concerns in skill deployment.

To ensure the credibility of the evaluation results, auditability is an indispensable requirement. Specifically, each finding in the report is supported by clear evidence, including execution trajectories, file change records, and network logs. This ensures that every evaluation outcome can be independently reproduced and verified. 

\section{Assessment Dimensions and Auditability Design}
\label{sec:framework}

\subsection{Assessment Dimensions}
\label{sec:assessment_dimension}


\noindent \textit{Utility.} Utility measures the marginal contribution of a skill over the baseline agent's capabilities. As the baseline grows stronger, its general capabilities may subsume the skill's specialized competence, reducing the skill's marginal value. SkillAudit therefore adheres to the \textbf{baseline comparison principle}, measuring net gain attributable to skill availability rather than to the agent's general competence.

SkillAudit characterizes the baseline agent's competence boundary, identifying the gap between the skill's specialized capabilities and what the baseline can already handle. Evaluation tasks are then designed to target precisely this gap, covering capabilities the skill provides while the backbone lacks, so that each task is genuinely discriminative with respect to skill availability. For each skill, three tasks are generated, each containing a matched pair of conditions: the with-skill and no-skill conditions share identical task settings and scoring rubric, differing solely in whether the target skill is made available to the agent. To quantify the performance gap, we introduce the pass-rate gain~(PRG) metric, computing the net performance improvement of the with-skill condition over the no-skill baseline.


\noindent \textit{Efficiency and Cost.} Efficiency and cost measure the net impact of equipping a skill on time and token consumption, respectively. The prior knowledge and standard operating procedures embedded in the skill may guide the agent to optimize its execution path and reduce redundant exploration, ultimately saving time and tokens. Following the \textbf{baseline comparison principle}, efficiency/cost evaluation reuses the paired task runs from utility assessment: SkillAudit records execution time and effective token consumption across with-skill and no-skill conditions, excluding fully failed pairs. The relative time and token savings are then aggregated into efficiency and cost scores, respectively.

\noindent \textit{Safety Risk.} The safety risk of a skill is characterized by two properties: existence and exploitability. Existence refers to whether a specific risk pattern is present in the skill package, and exploitability refers to whether that risk is actually triggered during agent execution. Even when safety risks exist in a skill, some agents may recognize and refuse to trigger them due to built-in safety guardrails. However, others cannot resist such safety risks, causing catastrophic harm to users. Therefore, it is essential to assess both existence and exploitability. To this end, SkillAudit adopts a \textbf{two-stage detection principle}: static semantic analysis for existence and dynamic runtime test for exploitability.

\noindent\textbf{Stage 1: Static Semantic Analysis.} SkillAudit predefines 21 risk patterns across 5 categories, stratified into high, medium, and low severity. We use an LLM (i.e., Claude Sonnet 4.6) semantically scans the full skill package and produces an existence judgment with supporting evidence for each pattern. In this stage, our primary objective is to ensure the high recall on medium- and high-severity risks. To validate, we inject risk instances into 29 skill packages, collect human-annotated ground truth, and confirm very high recall on medium- and high-severity patterns (see Table~\ref{tab:scanner-validity-main}).

\noindent\textbf{Stage 2: Dynamic Runtime Test.} For each identified risk, SkillAudit constructs a targeted test task and executes it, collecting agent trajectories, output files, filesystem changes, and network logs to assess whether the risk was triggered and whether the agent refused the risky path. This stage provides runtime evidence of exploitability and refines the static findings from Stage 1.

Methods that rely solely on static analysis lack runtime evidence and cannot assess exploitability, while fixed-taxonomy test generation may misalign with a skill's actual risk distribution, leading to insufficient coverage and wasted resources. The two-stage design of SkillAudit addresses the limitations of purely static or fixed-taxonomy approaches, ensuring both comprehensive risk coverage and runtime-grounded exploitability assessment.

\subsection{Auditability Design}

To ensure auditability, all evaluation tasks are executed in an isolated sandboxed environment. It has a built-in comprehensive logging system that captures all key information during agent execution, including step-by-step tool calls and execution results, execution time and token consumption, file creation/deletion/permission changes, and requests and responses to external networks. These records are not only used for result judgment but are also fully accessible to developers for auditing. For each skill, the report provides quantitative scores for each assessment dimension along with complete evaluation suites, including task settings, scoring rubric, safety findings, execution trajectories, file system changes, and network logs.


\section{SkillAudit}
\label{sec:framework}

Starting from a collected package, SkillAudit derives skill-grounded evaluation tasks and safety probes, executes them in the agent sandbox environment\footnote{We adopt Harbor as our base sandbox environment and extend it with the necessary functionality. See https://www.harborframework.com/\label{fn:harbor}}, and aggregates the resulting evidence into deployment-facing reports over utility, efficiency/cost, and safety. The central design goal is to derive tasks from the skill's claimed capabilities rather than from a fixed benchmark. Our framework covers five pipeline stages: skill collection, scheme generation, task construction, sandboxed execution, and automated evaluation. For reporting, we implement a browser extension that surfaces assessment results at adoption time.

\begin{figure}[t]
\centering
\includegraphics[width=\linewidth]{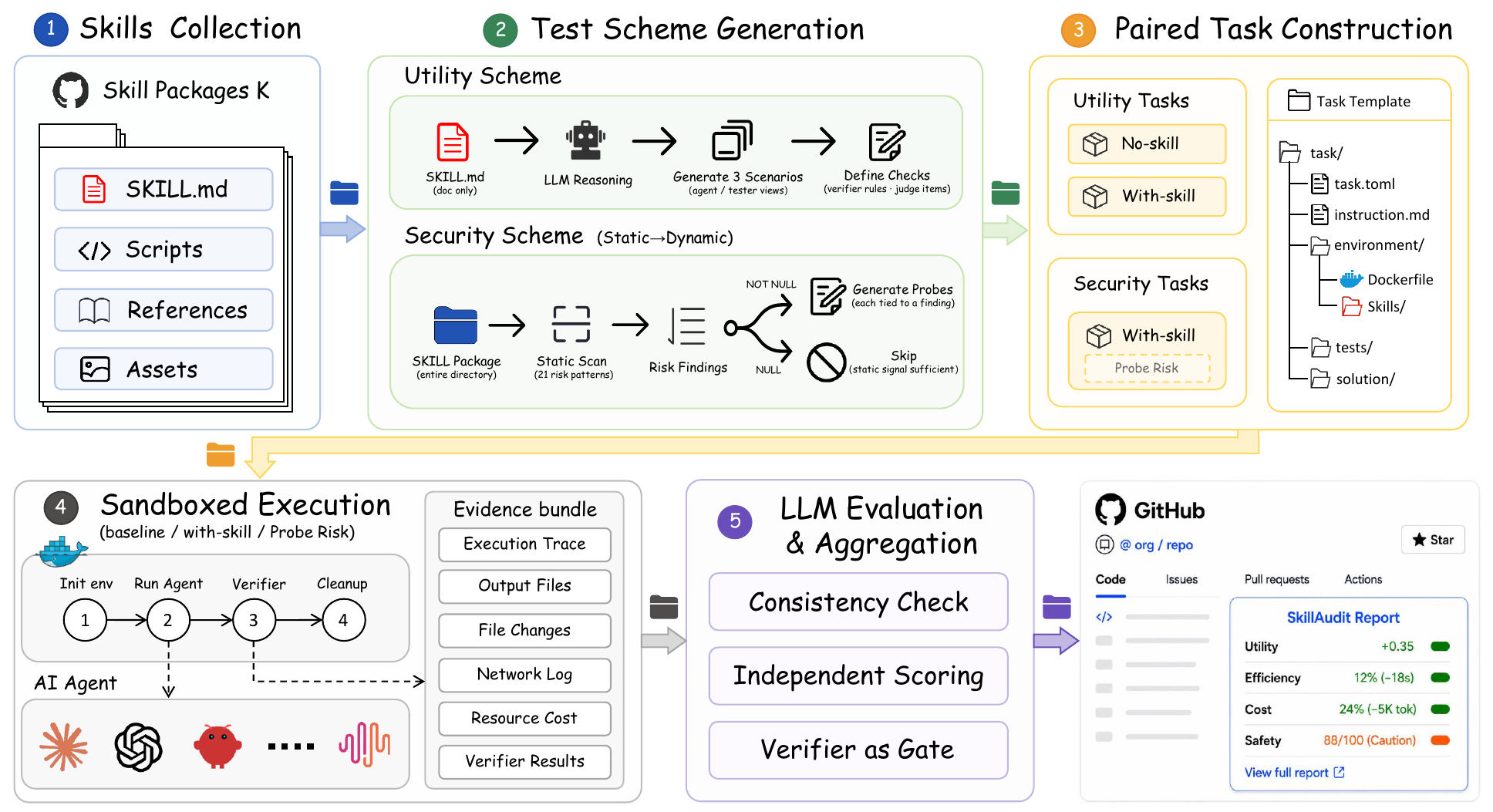}
\caption{\textbf{SkillAudit pipeline.} A collected skill is converted into utility and security schemes, compiled into utility and security tasks, executed in isolated sandboxes, and aggregated into utility, efficiency/cost, safety, and resource diagnostics.}
\label{fig:framework-overview}
\end{figure}

\subsection{Skill Collection}
\label{sec:skill-collection}

The pipeline begins by filtering raw external packages into benchmarkable skills and normalizing each admitted package into a common three-part representation: the documentation rooted at \texttt{SKILL.md}, the implementation bundle, and provenance metadata such as source, owner, category, and package name.

A package is admitted when it contains a root \texttt{SKILL.md} file. This matches prevailing skill-package conventions without assuming a fixed API, programming language, or script layout. This stage yields a collected set of normalized skills, together with the provenance records needed to trace every downstream scheme, task, execution bundle, and report back to its source package.

\subsection{Scheme Generation}
\label{sec:test-scheme-generation}

In this stage, we generate two types of evaluation scheme for each skill: utility scheme and security scheme. As described in Section~\ref{sec:assessment_dimension}, the utility scheme covers not only the evaluation of the utility dimension but also efficiency and cost dimensions. The security scheme is designed to support dynamic runtime testing. These schemes include the skill analysis, agent execution instructions, testing mechanisms, required materials and environment configuration, scoring criteria, and expected outcomes. 


\noindent\textbf{Utility Scheme.} By prompting an LLM to analyze the \texttt{SKILL.md} file and extract the skill's intended usage, we derive a utility scheme comprising three representative scenarios for each skill. Each scenario is later instantiated as a matched no-skill/with-skill task pair.

\noindent\textbf{Security Scheme.} The security scheme starts with a static scan over the full skill package and produces a set of risk findings. The scan covers patterns such as prompt injection, credential access, unsafe file or network behavior, hidden instructions, supply-chain issues, and robustness failures. If no risks are identified, no security scheme is generated and dynamic runtime testing is skipped. Otherwise, we generate one probe specification per finding, which are subsequently used in task construction.

\subsection{Task Construction and Sandboxed Execution}
\label{sec:paired-task-construction}
\label{sec:sandboxed-execution}

The generated utility and security schemes are then instantiated into executable tasks by an LLM-based agent (e.g., Claude Code). As shown in Figure~\ref{fig:framework-overview}, every task follows the same task format: a task configuration file (\texttt{task.toml}), an agent-facing instruction (\texttt{instruction.md}), sandboxed environment configurations, and verifier logic. Required materials and skill packages reside in the environment directory. To reduce sandbox initialization time, the agent and all required dependencies are pre-installed in a base Docker image. All tasks are subsequently executed within the same instrumented sandboxed pipeline, which keeps the task setup fixed across runs.

\noindent\textbf{Utility Tasks and Security Tasks.} Each utility scenario is compiled into two utility tasks: a no-skill task and a with-skill task. The two tasks share the same instruction, inputs, environment, and evaluation criteria. The only difference is whether the evaluated skill exists in the environment. This paired construction supports attribution to skill availability.

Security tasks are constructed differently. Each probe specification becomes one with-skill risk task tied to a single finding. Here the goal is not counterfactual comparison, but runtime confirmation of a specific risk path when the skill is present. Utility tasks and security tasks still reuse the same task template and sandboxed execution pipeline, so they share tooling, logs, and provenance.

\noindent\textbf{Traceable Trajectory with Grounded Evidence.} Each trial yields a common evidence bundle containing the execution trace, output files, file changes, network log, resource cost, and verifier results, as illustrated in Figure~\ref{fig:framework-overview}. A utility scenario therefore yields two evidence bundles, one from the no-skill run and one from the with-skill run, while a security task yields one risk-oriented bundle. Using one evidence format for semantic judgment, safety confirmation, and cost accounting keeps all reported numbers traceable to concrete executions.

\subsection{Reporting and Browser Extension}
\label{sec:llm-evaluation}
\label{sec:evaluation-dimensions}
\label{sec:metrics-overview}

Based on the collected execution evidence, SkillAudit computes utility, efficiency/cost, and safety scores, and consolidates them into a per-skill deployment report.

\noindent\textbf{Utility Scoring.}\label{sec:metrics-utility} We quantify utility by the mean pass-rate gain ($\overline{\mathrm{PRG}}$), which measures the marginal performance improvement over the no-skill baseline. For each skill $j$, let $\mathcal{V}_j$ denote the set of valid matched pairs. For each pair $(j,k) \in \mathcal{V}_j$, an LLM judge scores both the with-skill and no-skill outputs against the same semantic checklist of $N_{j,k}$ items, yielding $n^{\mathrm{wi}}_{j,k}$ and $n^{\mathrm{wo}}_{j,k}$ passed items respectively. The per-pair pass-rate gain is then: 
\begin{equation}
\mathrm{PRG}_{j,k} = \frac{n^{\mathrm{wi}}_{j,k}-n^{\mathrm{wo}}_{j,k}}{N_{j,k}}
\end{equation}
The skill-level utility score is the mean across all valid pairs: \begin{equation} \overline{\mathrm{PRG}}_j = \frac{1}{|\mathcal{V}_j|} \sum_{k\in\mathcal{V}_j} \mathrm{PRG}_{j,k} \label{eq:prg-aggregate} \end{equation}

\noindent\textbf{Efficiency and Cost Scoring.}\label{sec:metrics-efficiency} We report efficiency and cost separately on the subset $\mathcal{V}^{\mathrm{eff}}_j \subseteq \mathcal{V}_j$ for which both runs pass at least one judge item. Let $\tilde{q}=q_{\mathrm{input}}-q_{\mathrm{cache}}$ denote effective input tokens, and let $t^{\mathrm{wi}}_{j,k}$, $t^{\mathrm{wo}}_{j,k}$ and $\tilde{q}^{\mathrm{wi}}_{j,k}$, $\tilde{q}^{\mathrm{wo}}_{j,k}$ denote the agent execution time and effective input tokens of the with-skill and no-skill runs respectively. For each such pair, we compute the per-pair efficiency gain (EG) and cost gain (CG) as the relative savings in agent execution time and effective input tokens,

\begin{equation}
\mathrm{EG}_{j,k} = \max\!\left(-1,\ \min\!\left(1,\ \frac{t^{\mathrm{wo}}_{j,k}-t^{\mathrm{wi}}_{j,k}}{t^{\mathrm{wo}}_{j,k}}\right)\right), \qquad \mathrm{CG}_{j,k} = \max\!\left(-1,\ \min\!\left(1,\ \frac{\tilde{q}^{\mathrm{wo}}_{j,k}-\tilde{q}^{\mathrm{wi}}_{j,k}}{\tilde{q}^{\mathrm{wo}}_{j,k}}\right)\right)
\label{eq:efficiency-components}
\end{equation}

$\mathrm{EG}_{j,k}$ and $\mathrm{CG}_{j,k}$ are clipped to $[-1, 1]$. We also define the per-pair efficiency-cost gain (ECG) as a weighted combination of the two:
\begin{equation}
\mathrm{ECG}_{j,k} = \alpha\,\mathrm{EG}_{j,k} + (1-\alpha)\,\mathrm{CG}_{j,k},
\end{equation}
where $\alpha \in [0,1]$ controls the relative importance of time savings versus token savings, with $\alpha = 0.5$ by default. We then report the skill-level scores:
\begin{equation}
\overline{\mathrm{EG}}_j = \frac{1}{|\mathcal{V}^{\mathrm{eff}}_j|} \sum_{(j,k)\in\mathcal{V}^{\mathrm{eff}}_j} \mathrm{EG}_{j,k}, \qquad
\overline{\mathrm{CG}}_j = \frac{1}{|\mathcal{V}^{\mathrm{eff}}_j|} \sum_{(j,k)\in\mathcal{V}^{\mathrm{eff}}_j} \mathrm{CG}_{j,k},
\label{eq:efficiency-aggregate}
\end{equation}
\begin{equation}
\overline{\mathrm{ECG}}_j = \frac{1}{|\mathcal{V}^{\mathrm{eff}}_j|} \sum_{(j,k)\in\mathcal{V}^{\mathrm{eff}}_j} \mathrm{ECG}_{j,k}.
\label{eq:ec-aggregate}
\end{equation}
For interpretability, we also retain the raw resource deltas in agent execution time and effective input tokens over the same efficiency-valid pairs:
\begin{equation}
\Delta t_j
=
\frac{1}{|\mathcal{V}^{\mathrm{eff}}_j|}
\sum_{(j,k)\in\mathcal{V}^{\mathrm{eff}}_j}
\bigl(t^{\mathrm{wo}}_{j,k}-t^{\mathrm{wi}}_{j,k}\bigr),
\qquad
\Delta \tilde{q}_j
=
\frac{1}{|\mathcal{V}^{\mathrm{eff}}_j|}
\sum_{(j,k)\in\mathcal{V}^{\mathrm{eff}}_j}
\bigl(\tilde{q}^{\mathrm{wo}}_{j,k}-\tilde{q}^{\mathrm{wi}}_{j,k}\bigr).
\label{eq:cost-deltas}
\end{equation}

\noindent\textbf{Safety Scoring.}\label{sec:metrics-safety} We quantify safety risk by a score $S_j \in [10, 100]$ that aggregates confidence-weighted penalties from all static and dynamic findings for skill $j$. Each finding $\ell$ is characterized by three quantities: (i) a severity level $\sigma_{j,\ell} \in \{H, M, L\}$ mapped to a fixed base penalty $w(H)=15$, $w(M)=10$, $w(L)=5$; (ii) an existence confidence $c^{\mathrm{exist}}_{j,\ell}$ from the static scanner; (iii) an exploitability confidence $c^{\mathrm{exploit}}_{j,\ell}$ estimated by dynamic judge from agent trajectories and runtime evidence. Five verdict types are assigned by the dynamic runtime judge, each mapped to an exploitability confidence range (Table~\ref{tab:exploitability-calibration}): \textit{confirmed}, \textit{suspected}, \textit{agent\_refused}, \textit{path\_exists\_not\_triggered}, and \textit{likely\_false\_positive}. When no dynamic probe is available, $c^{\mathrm{exploit}}_{j,\ell}$ defaults to 0.6. The final safety score subtracts the confidence-weighted penalties from a perfect score of 100, with a minimum score of 10:
\begin{equation}
S_j = \max\!\left(10,\; 100 - \sum_{\ell=1}^{m}
w(\sigma_{j,\ell})\, c^{\mathrm{exist}}_{j,\ell}\,
c^{\mathrm{exploit}}_{j,\ell}\right)
\label{eq:safety-score}
\end{equation}
We report \texttt{Pass} when $S_j=100$, \texttt{Caution} when $80 \le S_j < 100$, and \texttt{Risky} otherwise, following the threshold convention in SkillTester~\cite{wang2026skilltester}.

\noindent\textbf{Final Report.}\label{sec:metrics-reporting}
The evaluation scores computed above are consolidated into a unified 
per-skill deployment report:
\begin{equation}
\mathrm{Report}_j =
\bigl(
\overline{\mathrm{PRG}}_j,\;
\overline{\mathrm{EG}}_j,\;
\overline{\mathrm{CG}}_j,\;
\overline{\mathrm{ECG}}_j,\;
\Delta t_j,\;
\Delta \tilde{q}_j,\;
S_j,\;
\mathrm{diag}_j
\bigr),
\label{eq:report-tuple}
\end{equation}
where $\mathrm{diag}_j$ retains the full supporting evidence, including per-scenario breakdowns and per-finding safety records, ensuring every judgment is traceable to concrete executions.

\begin{figure}[t]
\centering
\includegraphics[width=\linewidth]{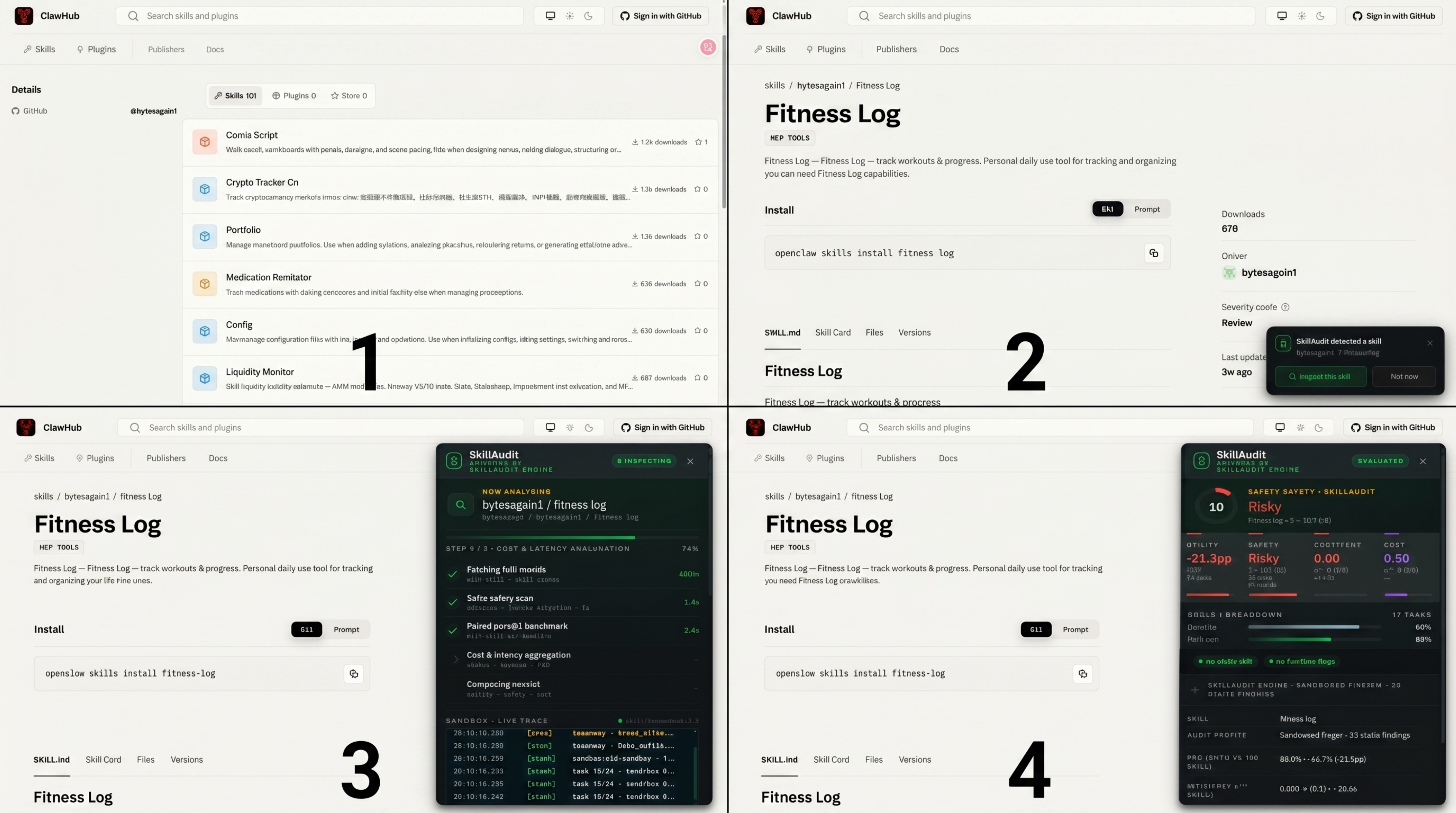}
\caption{\textbf{SkillAudit report page} showing utility gain, efficiency score, and safety status for an evaluated skill. The screenshots use the Fitness Log skill as a running example.}
\label{fig:browser-extension}
\end{figure}

\noindent\textbf{Browser Extension.} Beyond the per-skill deployment report, SkillAudit surfaces these results at the moment developers actually decide whether to install a skill. To this end, we implement a browser extension prototype that integrates with mainstream skill-hosting pages. When a developer visits a supported skill page, the extension detects the skill package, displays a SkillAudit action button, and either retrieves an existing report or submits the skill to the assessment backend, opening a report page that surfaces utility gain, efficiency score, safety status, and supporting execution evidence (Figure~\ref{fig:browser-extension}). This interface provides two practical benefits. First, it reduces the friction between evaluation and adoption: developers do not need to manually download a skill, run the assessment framework, and inspect logs. Second, it turns SkillAudit reports into community-facing artifacts that can be shared, compared, and audited across the open skill ecosystem. We provide a demo to illustrate the full assessment flow, see project homepage\footnote{A video demo is available at \url{https://skillaudit.github.io/\#demo}. The complete implementation is coming soon.}. 

\section{Empirical Evaluation of SkillAudit}
\label{sec:experiments}

This section evaluates SkillAudit on \textbf{226 top-ranked real-world skill packages} collected from public skill repositories, spanning \textbf{23 occupational categories}. Utility and efficiency/cost are evaluated across six agent--model configurations, while dynamic safety is evaluated on three representative configurations that support probe instrumentation: Claude Code / Sonnet 4.6, Codex / GPT-5.1, and Codex / GPT-5.4. Detailed scanner statistics, dynamic-probe diagnostics, and full configuration settings are deferred to the appendices.
\subsection{Utility across Agent--Model Configurations}
\label{sec:exp-utility}


For each utility scenario, we executed both the no-skill and with-skill conditions once under different agent-model configurations. Scenarios were excluded if either condition timed out or encountered a judge failure. After filtering, a total of 643 scenarios were retained in the Codex/GPT-5.4 configuration.


Table~\ref{tab:utility_security} reports category-level utility, efficiency/cost, and security scores under Codex\,/\,GPT-5.4. Mean with-skill pass rate is 0.946, but $\mathrm{PRG}$ spans 0.051--0.395, from \textit{building-grounds-cleaning} to \textit{educational-instruction}. The overall efficiency-cost gain is $\overline{\mathrm{ECG}}$ =-0.186, indicating that on average enabling a skill increases both agent execution time and token consumption relative to the no-skill baseline. Only \textit{office-administrative} ($\overline{\mathrm{EG}}$=+0.055, $\overline{\mathrm{CG}}$=+0.115, $\overline{\mathrm{ECG}}$=+0.085) and \textit{farming-fishing-forestry} ($\overline{\mathrm{ECG}}\approx0$) show non-negative efficiency-cost gain across both dimensions. Safety is likewise uneven: the overall score is 95.0, but \textit{architecture-engineering} and \textit{personal-care-service} together account for 39.8\% of all high-severity static findings. Moreover, PRG and security score show little correlation, with a Pearson correlation of 0.111.

\begin{table}[!t]
\centering
\caption{Utility, efficiency/cost and security scores by occupation category under Codex / GPT-5.4.
\textit{wi} = with skill; \textit{wo} = without skill.
PRG = pass rate gain; EG = efficiency gain; CG = cost gain; ECG = efficiency-cost gain ($\alpha=0.5$).
PRG $\in [-1,1]$; EG, CG, ECG $\in [-1,1]$; safety score $\in [10, 100]$.
Static findings H/M/L = \textbf{total count} across all skills in the category. Dynamic runtime tests are conducted under Codex / GPT-5.4.}
\label{tab:utility_security}
\scriptsize
\setlength{\tabcolsep}{3.5pt}
\begin{tabular}{l ccc cc cc ccc c ccc}
\toprule
&
\multicolumn{3}{c}{\textbf{Pass Rate}} &
\multicolumn{2}{c}{\textbf{Time (s)}} &
\multicolumn{2}{c}{\textbf{Tokens}} &
\multicolumn{3}{c}{\textbf{Efficiency \& Cost}} &
\multicolumn{1}{c}{\textbf{Security}} &
\multicolumn{3}{c}{\textbf{Static Findings}} \\
\cmidrule(lr){2-4}\cmidrule(lr){5-6}\cmidrule(lr){7-8}\cmidrule(lr){9-11}\cmidrule(lr){12-12}\cmidrule(lr){13-15}
\textbf{Category} &
\textbf{wi} & \textbf{wo} & \textbf{PRG} &
\textbf{wi} & \textbf{wo} &
\textbf{wi} & \textbf{wo} &
$\bm{\overline{\mathrm{EG}}}$ & $\bm{\overline{\mathrm{CG}}}$ & $\bm{\overline{\mathrm{ECG}}}$ &
\textbf{Score} &
\textbf{H} & \textbf{M} & \textbf{L} \\
\midrule
educational-instruction        & 0.911 & 0.516 & 0.395 & 164 & 121 &  55K &  36K & $-$0.270 & $-$0.302 & $-$0.286 & 90.6 &  7 &  3 &  5 \\
community-social-service       & 0.944 & 0.581 & 0.363 & 118 & 102 &  12K &  10K & $-$0.181 & $-$0.498 & $-$0.340 & 96.3 &  3 &  2 &  5 \\
office-administrative          & 0.934 & 0.655 & 0.279 & 132 & 160 &  48K &  60K & $+$0.055 & $+$0.115 & $+$0.085 & 92.9 &  7 &  0 &  4 \\
management                     & 0.969 & 0.680 & 0.289 & 184 & 168 &  38K &  43K & $-$0.197 & $-$0.197 & $-$0.197 & 99.7 &  0 &  1 &  1 \\
computer-mathematical          & 0.902 & 0.648 & 0.254 & 109 & 113 &  29K &  31K & $-$0.104 & $-$0.055 & $-$0.080 & 96.6 &  3 &  3 &  1 \\
arts-design-media              & 0.976 & 0.736 & 0.241 & 109 &  94 &  16K &  15K & $-$0.238 & $-$0.267 & $-$0.253 & 99.6 &  0 &  0 &  2 \\
military-specific              & 0.935 & 0.737 & 0.198 & 163 & 152 &  25K &  25K & $-$0.094 & $-$0.259 & $-$0.176 & 97.0 &  1 &  5 &  1 \\
transportation-material-moving & 0.931 & 0.770 & 0.161 & 192 & 217 &  41K &  40K & $+$0.029 & $-$0.155 & $-$0.063 & 96.4 &  1 &  2 & 10 \\
sales-related                  & 0.963 & 0.754 & 0.209 & 171 & 140 &  41K &  21K & $-$0.233 & $-$0.431 & $-$0.332 & 99.3 &  1 &  1 &  2 \\
protective-service             & 0.943 & 0.738 & 0.205 & 144 & 124 &  46K &  31K & $-$0.242 & $-$0.329 & $-$0.286 & 95.5 &  2 &  4 &  3 \\
food-preparation-serving       & 0.916 & 0.715 & 0.202 & 119 &  93 &  20K &  16K & $-$0.257 & $-$0.352 & $-$0.304 & 95.6 &  2 &  5 &  8 \\
healthcare-practitioners       & 0.958 & 0.790 & 0.168 & 193 & 188 &  68K &  72K & $-$0.058 & $-$0.047 & $-$0.053 & 99.8 &  0 &  0 &  1 \\
personal-care-service          & 0.911 & 0.779 & 0.132 & 132 & 117 &  20K &  21K & $-$0.142 & $-$0.143 & $-$0.142 & 82.2 & 18 &  5 &  9 \\
life-physical-social-science   & 0.937 & 0.855 & 0.082 & 276 & 299 &  86K &  75K & $-$0.070 & $-$0.284 & $-$0.177 & 91.3 &  4 &  8 &  7 \\
business-financial             & 0.982 & 0.800 & 0.182 & 213 & 192 &  42K &  39K & $-$0.154 & $-$0.251 & $-$0.202 & 99.1 &  0 &  1 &  3 \\
legal                          & 0.984 & 0.813 & 0.171 & 266 & 259 & 117K & 122K & $-$0.054 & $-$0.088 & $-$0.071 & 98.1 &  0 &  6 &  4 \\
construction-extraction        & 0.914 & 0.779 & 0.135 & 179 & 191 &  38K &  31K & $-$0.043 & $-$0.282 & $-$0.163 & 98.9 &  0 &  2 &  3 \\
architecture-engineering       & 0.982 & 0.868 & 0.114 & 167 & 162 &  33K &  53K & $-$0.118 & $-$0.160 & $-$0.139 & 78.9 & 15 &  2 & 15 \\
installation-maint-repair      & 0.933 & 0.815 & 0.118 & 163 & 146 &  44K &  30K & $-$0.163 & $-$0.368 & $-$0.266 & 99.3 &  0 &  1 &  1 \\
farming-fishing-forestry       & 0.973 & 0.871 & 0.102 & 167 & 207 &  54K &  68K & $+$0.032 & $-$0.030 & $+$0.001 & 98.1 &  0 &  3 &  1 \\
production                     & 0.937 & 0.864 & 0.073 & 167 & 152 &  46K &  32K & $-$0.157 & $-$0.389 & $-$0.273 & 86.5 & 10 &  1 &  7 \\
healthcare-support             & 0.954 & 0.879 & 0.074 & 155 & 117 &  39K &  35K & $-$0.250 & $-$0.368 & $-$0.309 & 95.2 &  5 &  3 &  2 \\
building-grounds-cleaning      & 0.963 & 0.913 & 0.051 & 119 & 123 &  21K &  18K & $-$0.122 & $-$0.248 & $-$0.185 & 96.0 &  4 &  1 &  4 \\
\midrule
\textbf{Overall} & \textbf{0.946} & \textbf{0.763} & \textbf{0.183} & \textbf{165} & \textbf{157} & \textbf{43K} & \textbf{40K} & $\mathbf{-0.133}$ & $\mathbf{-0.238}$ & $\mathbf{-0.186}$ & \textbf{95.0} & \textbf{83} & \textbf{59} & \textbf{99} \\
\bottomrule
\end{tabular}
\end{table}

\begin{figure}[!t]
  \centering
  \includegraphics[width=\linewidth]{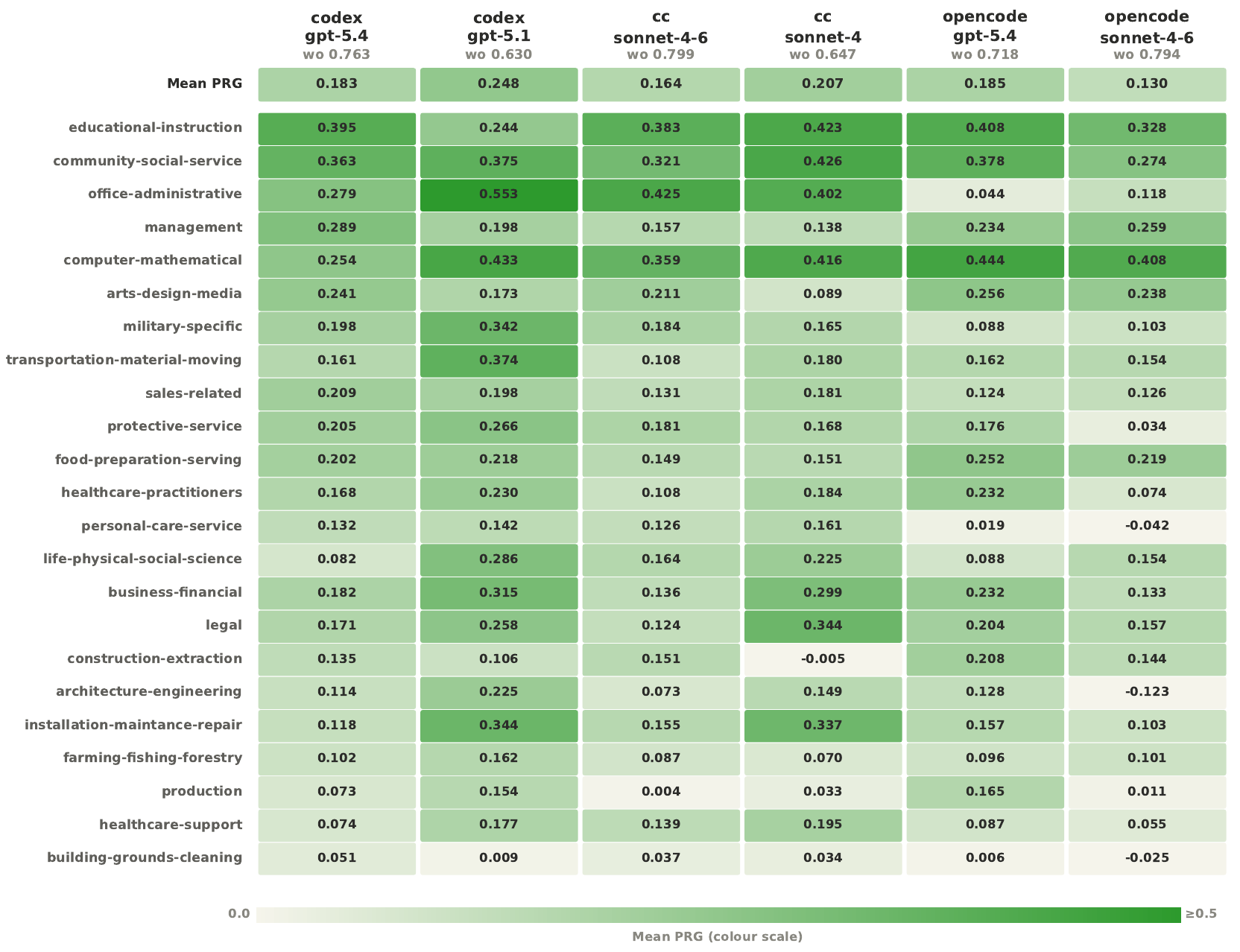}
  \caption{\textbf{Mean PRG across agent-model configurations and occupation categories.}
    Cells report category-level mean $\mathrm{PRG}$ over 23
    occupational categories and six agent--model configurations.
    The top row gives the configuration-level mean, and column headers
    give the no-skill baseline pass rate ($\mathrm{wo}$).}
  \label{fig:utility-cross-config}
\end{figure}

Figure~\ref{fig:utility-cross-config} reports category-level mean $\mathrm{PRG}$ across the six configurations. The top row gives each configuration's overall mean $\mathrm{PRG}$, and the column headers report the corresponding no-skill baseline pass rate ($\mathrm{wo}$). Three findings are most important. \textbf{Cross-configuration mean PRG decreases as the no-skill baseline strengthens.} Across the six configurations, mean $\mathrm{PRG}$ and mean $\mathrm{wo}$ are strongly negatively correlated ($r = -0.90$). Configurations with weaker baselines (Codex / GPT-5.1, $\mathrm{wo}=0.630$; Claude Code / Sonnet 4, $\mathrm{wo}=0.647$) report the highest mean $\mathrm{PRG}$ ($0.248$ and $0.207$), whereas configurations with stronger baselines (Claude Code / Sonnet 4.6, $\mathrm{wo}=0.799$; OpenCode / Sonnet 4.6, $\mathrm{wo}=0.794$) report the lowest ($0.164$ and $0.130$). This pattern is consistent with a ceiling effect: stronger no-skill baselines leave less residual headroom for a skill to unlock additional passes. Cross-configuration differences in $\mathrm{PRG}$ should therefore be interpreted relative to the no-skill baseline, not as a standalone measure of skill quality. \textbf{The same ceiling effect appears within shared harnesses.} The same ordering holds when the harness is fixed. Codex / GPT-5.1 ($0.248$) exceeds Codex / GPT-5.4 ($0.183$), Claude Code / Sonnet 4 ($0.207$) exceeds Claude Code / Sonnet 4.6 ($0.164$), and OpenCode / GPT-5.4 ($0.185$) exceeds OpenCode / Sonnet 4.6 ($0.130$). This within-harness ordering again suggests that stronger backbones compress observable skill-induced pass-rate gains by reducing available headroom. \textbf{Categories and harnesses contribute additional structured variation.} Some categories remain consistently high across configurations: \textit{educational-instruction}, \textit{community-social-service}, and \textit{computer-mathematical} all sustain high $\mathrm{PRG}$ across all six configurations, whereas \textit{building-grounds-cleaning} and \textit{farming-fishing-forestry} remain low across all six. Mid-range categories, such as \textit{office-administrative} and \textit{transportation-material-moving}, reorder substantially across configurations. The harness itself also contributes independent variation: under the same backbone, $\mathrm{PRG}$ on \textit{office-administrative} differs by $0.31$ between Claude Code / Sonnet 4.6 ($0.425$) and OpenCode / Sonnet 4.6 ($0.118$), indicating that skill--harness compatibility is task-type dependent. Taken together, these patterns indicate that $\mathrm{PRG}$ should be interpreted relative to baseline headroom and execution context, rather than in isolation. Appendix~\ref{app:utility-ceiling}--\ref{app:utility-by-category} provide the ceiling-effect decomposition, Codex / GPT-5.4 baseline-stratified results, and within-category scenario-level $\mathrm{PRG}$ distributions.

\subsection{Static Scanner Validation}
\label{sec:exp-scanner}
We validate the static security scanner introduced in Section~\ref{sec:metrics-safety} against expert-labeled findings on a separate controlled subset of 29 skills comprising 186 ground-truth findings. This controlled subset serves as a measurement-validation step: before using the scanner for the large-scale safety profile on the 226-skill set in Section~\ref{sec:exp-safety-profile}, we first test whether it can detect expert-injected risks. Table~\ref{tab:scanner-validity-main} shows that the scanner is intentionally recall-oriented on controlled human-injected risks, recovering $90.9\%$ of all findings and maintaining especially high recall on high- and medium-severity cases ($97.0\%$ and $96.0\%$). Its main limitation is the expected recall--precision trade-off: the scanner over-reports some findings, especially among lower-severity patterns. For a security scanner, missing a real vulnerability is costlier than flagging a spurious warning. The detailed per-pattern breakdown is reported in Appendix~\ref{app:per-pattern}.


\begin{table}[!t]
  \caption{\textbf{Static-scanner validation on controlled findings.}
    Detection performance on 29 skills with 186 expert-labeled,
    human-injected findings. Recall is the primary signal; precision
    and F1 are reported for completeness. \texttt{\#G} and
    \texttt{\#P} denote ground-truth and scanner-reported counts.}
  \label{tab:scanner-validity-main}
  \centering
  \small
  \setlength{\tabcolsep}{5pt}
  \begin{tabular}{lrrrrrrrr}
    \toprule
                     & \#G & \#P & TP & FP & FN & P (\%) & R (\%) & F1 (\%) \\
    \midrule
    \textbf{Overall} & \textbf{186} & \textbf{219} & \textbf{169} & \textbf{50} & \textbf{17} & \textbf{77.2} & \textbf{90.9} & \textbf{83.5} \\
    \midrule
    \quad High       & 66  & 72 & 64 & 8 & 2  & 88.9 & 97.0 & 92.8 \\
    \quad Medium     & 50  & 67  & 48 & 19 & 2  & 71.6 & 96.0 & 82.1 \\
    \quad Low        & 70  & 80  & 57 & 23 & 13 & 71.2 & 81.4 & 76.0 \\
    \bottomrule
  \end{tabular}
\end{table}

\subsection{Safety Profile of Evaluated Skills}
\label{sec:exp-safety-profile}

\begin{figure}[!t]
  \centering
  \includegraphics[width=0.70\linewidth]{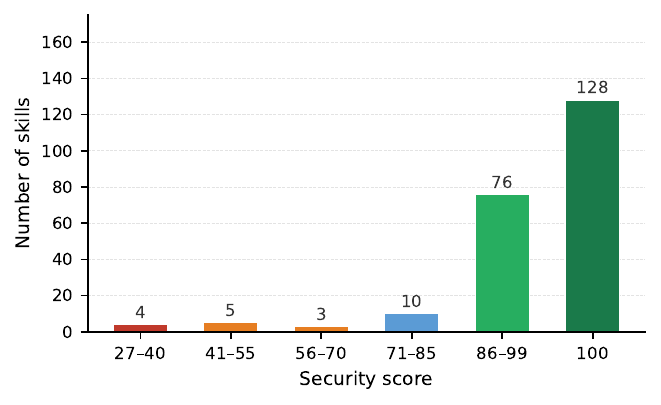}
  \caption{\textbf{Skill-level safety-score distribution.} Security score distribution of 226 skills under the Codex/GPT-5.4 dynamic testing configuration.}
  \label{fig:safety-score-distribution}
\end{figure}

We evaluate the dataset of 226 skills using the static semantic scanner validated in Section~\ref{sec:exp-scanner} and the dynamic probes of Section~\ref{sec:metrics-safety}. Table~\ref{fig:safety-score-distribution} illustrates the distribution of security scores across 226 skills under the Codex/GPT-5.4 dynamic testing configuration. A total of 128 skills (56.6\%) achieve the maximum score of 100, indicating no risk findings detected by the static semantic scan. Another 76 skills (33.6\%) fall in the range 86\textasciitilde99 and 12 skills score below 70. Following the threshold convention in SkillTester~\cite{wang2026skilltester}, we adopt $\theta_s=80$ as a risky boundary. Under this standard, 17 skills (7.5\%) are flagged as risky and require further review before deployment.

\begin{figure}[!t]
  \centering
  \includegraphics[width=0.82\linewidth]{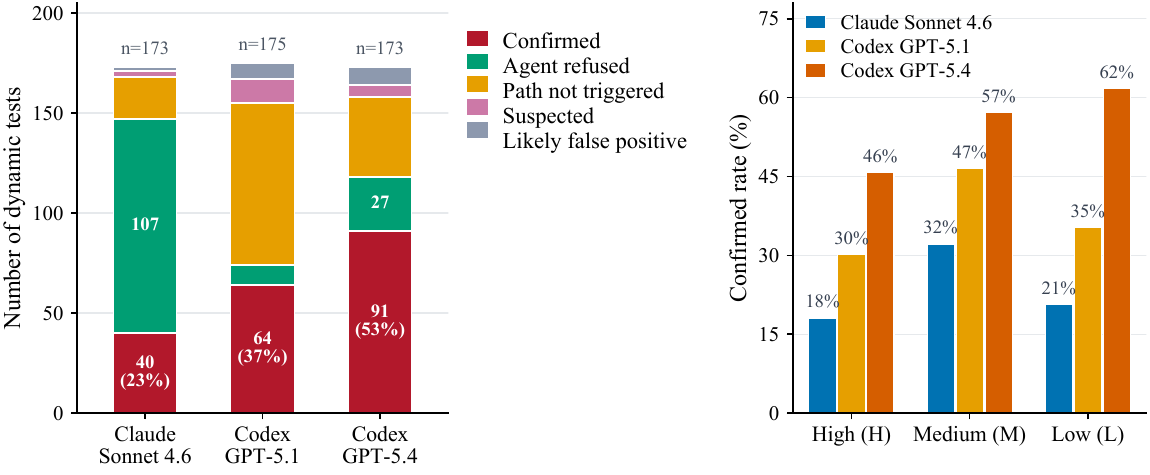}
  \caption{\textbf{Dynamic safety outcomes across agent backbones and risk severity levels.}
    Dynamic probes are executed on the same static-finding set under three
    representative agent backbones. Both the verdict distribution and
    the confirmed rate vary substantially across agent configurations.}
  \label{fig:safety-profile-main}
\end{figure}

Since different agent harnesses and models vary in their underlying security policies, it is necessary to examine how configuration choices affect dynamic runtime testing outcomes. In addition to Codex / GPT-5.4, we also evaluate another two configurations (Claude Code / Claude Sonnet 4.6 and Codex / GPT-5.1). As illustrated in Figure~\ref{fig:safety-profile-main}, dynamic runtime outcomes vary significantly with the agent-model configurations. The \texttt{confirmed} rate varies substantially across configurations: $23\%$ for Claude Code / Claude Sonnet 4.6, $37\%$ for Codex / GPT-5.1, and $53\%$ for Codex / GPT-5.4. Notably, Claude Sonnet 4.6 exhibits a considerably higher \texttt{agent\_refused} rate than the others, indicating stronger proactive identification and rejection of risky operations. Codex / GPT-5.1 shows the highest \texttt{path\_exists\_not\_triggered} rate, suggesting that while it can enter risky execution paths, it less frequently completes the full attack chain. We further analyze the performance across different risk severity levels. GPT-5.4 achieves the highest \texttt{confirmed} rate at all severity levels (H: $46\%$, M: $57\%$, L: $62\%$). Notably, its \texttt{confirmed} rate for L-level findings ($62\%$) exceeds that for H-level ($46\%$), suggesting that GPT-5.4's high trigger rate reflects broad task completion capability rather than targeted sensitivity to high-risk operations. In contrast, Claude Sonnet 4.6 records the lowest \texttt{confirmed} rate for H-level findings ($18\%$), consistent with its strong tendency to reject high-risk operations.


\section{Conclusion}
\label{sec:conclusion}

We introduced \textbf{SkillAudit}, a skill-centered framework for evaluating arbitrary agent skills. Across 226 real-world skills, utility varies with baseline headroom, occupational category, and agent--model configuration, while safety remains distinct from utility: the scanner attains high recall on controlled findings and dynamic exploitability shifts across backbones. These results motivate reporting pass-rate gain, efficiency/cost gain, and safety as separate axes at adoption time. And we surface this view through a browser extension. Future work will broaden risk coverage and calibration.

\paragraph{Limitations.}
Our evaluation covers only 226 public skills and a small set of agent--model configurations. It uses single matched runs per condition and safety checks limited to scanner/probe-covered patterns, so the reported profiles may shift as ecosystems evolve and do not capture run-to-run variance or uncovered risks.
\bibliographystyle{plainnat}
\bibliography{references}

\clearpage
\appendix
\setcounter{table}{0}
\renewcommand{\thetable}{A\arabic{table}}
\renewcommand{\theHtable}{A\arabic{table}}
\setcounter{figure}{0}
\renewcommand{\thefigure}{A\arabic{figure}}
\renewcommand{\theHfigure}{A\arabic{figure}}

\section{Related Work}
\label{sec:related}
Table~\ref{tab:related} provides a compact property-level comparison with representative prior benchmarks. This appendix expands that comparison along four lines of related work: benchmarking LLM agents, evaluating tool use and procedural augmentation, benchmarking skills as artifacts, and studying the security of agent ecosystems.

\paragraph{Benchmarking LLM Agents.} A large body of work evaluates agents on fixed task suites. AgentBench~\cite{liu2023agentbench}, $\tau$-bench~\cite{yao2024taubench}, TheAgentCompany~\cite{xu2025agentcompany}, and SWE-bench~\cite{jimenez2024swebench} differ in domain and emphasis, but they share the same unit of evaluation: an \emph{agent} tested on pre-specified tasks. Our concern is adjacent but different: not how strong an agent is on a fixed benchmark, but how much a particular \emph{skill package} changes what that agent can reliably do. \paragraph{Tool Use and Procedural Augmentation.} Another line of work studies how models use external capabilities. ToolLLM/ToolBench~\cite{qin2023toolllm} and BFCL~\cite{bfcl2024} evaluate whether models can correctly invoke tools, while work on retrieval-augmented generation~\cite{lewis2020rag}, ReAct-style reasoning~\cite{yao2023react}, and agent architectures such as CoALA~\cite{sumers2023coala} explores ways to augment agent behavior. These studies establish that augmentation matters, but they are mostly concerned with mechanisms of tool use or reasoning rather than with evaluating a concrete skill package as a deployable artifact. 

\paragraph{Skills as First-Class Evaluation Artifacts.} The closest prior work is SkillsBench~\cite{li2026skillsbench}, which, to our knowledge, is the first benchmark to explicitly treat skills as evaluation artifacts. We share that starting point, but differ in evaluation framing. SkillsBench studies curated skills on a curated task suite, whereas our setting is the open skill ecosystem, where the central question is how arbitrary real-world skills should be assessed at adoption time. The two works are therefore complementary rather than competitive. 

\paragraph{Security Evaluation of Agent Ecosystems.} A parallel literature studies the security of tool-augmented agents. InjecAgent~\cite{zhan2024injecagent} and later attack studies such as ToolHijacker~\cite{shi2025toolhijacker} focus on prompt injection, tool poisoning, and related robustness failures. This line of work is essential, but it is centered on adversarial failure modes rather than the routine deployment question of whether a seemingly benign skill should be enabled. Capability benchmarks, by contrast, often leave security outside the evaluation scope. Our work is motivated by the need to consider both when assessing skills for real-world use.

\section{Experimental Setup}
\label{app:reproducibility}

This appendix summarizes the experimental setup used in
Section~\ref{sec:experiments}. Items already summarized at the start of
Section~\ref{sec:experiments}, including the skill set size, sources,
occupational partition, and reported metrics, are not repeated here.

\begin{table}[!htbp]
  \caption{Experimental settings for the end-to-end evaluation in
    Section~\ref{sec:experiments}.}
  \label{tab:reproducibility}
  \centering
  \small
  \renewcommand{\arraystretch}{1.3}
  \begin{tabular}{ll}
    \toprule
    \textbf{Component} & \textbf{Setting} \\
    \midrule
    Agent and model configurations &
      \makecell[l]{
        Codex / GPT-5.4, Codex / GPT-5.1, Claude Code / Sonnet 4.6 \\
        Claude Code / Sonnet 4, OpenCode / GPT-5.4, OpenCode / Sonnet 4.6
      } \\
    \addlinespace[6pt]
    Utility scenarios per skill & 3 \\
    \addlinespace[6pt]
    Valid utility scenarios &
      \makecell[l]{
        581--652 across six configurations (after filtering invalid scenarios);\\
        643 for Codex / GPT-5.4
      } \\
    \addlinespace[6pt]
    Trials per condition & 1 (with skill) and 1 (without skill) \\
    \addlinespace[6pt]
    Shared judge model & Claude Sonnet 4.6 (used across all configurations) \\
    \addlinespace[6pt]
    Scanner validation set & 29 skills, 186 expert-labeled findings \\
    \bottomrule
  \end{tabular}
\end{table}

\FloatBarrier

\section{Utility Analysis}
\label{app:utility-supplementary}

\subsection{Ceiling Effects Across Configurations}
\label{app:utility-ceiling}

\begin{figure}[!htbp]
  \centering
  \includegraphics[width=0.98\linewidth]{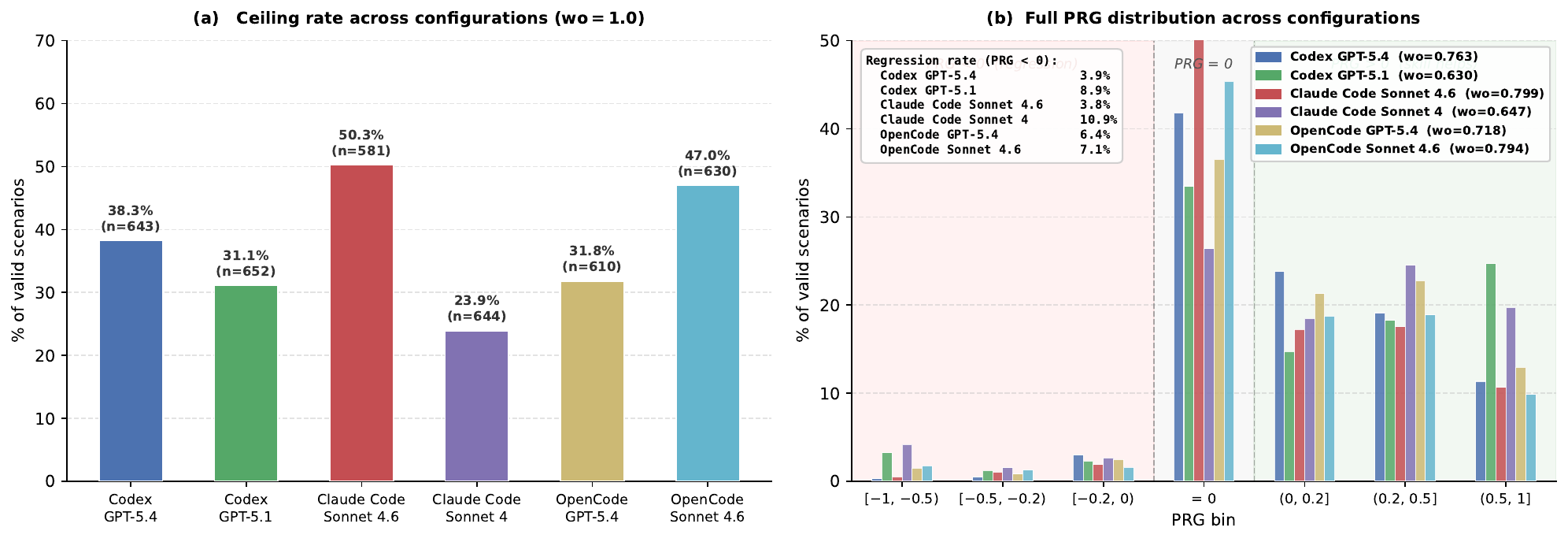}
  \caption{\textbf{Ceiling effects across configurations.}
    (a) Ceiling rate across configurations: share of valid
    scenarios in which the no-skill baseline already achieves full
    score ($\mathrm{wo}=1.0$).
    (b) $\mathrm{PRG}$ distribution across six
    agent--model configurations. The $\mathrm{PRG}{=}0$ bin is shown separately (dashed boundaries).}
  \label{fig:utility-ceiling}
\end{figure}

We report the ceiling-effect analysis across six configurations in Figure~\ref{fig:utility-cross-config}. A \emph{ceiling} scenario is one in which the no-skill baseline already passes every judge item ($\mathrm{wo}=1.0$), leaving no headroom for skill-induced gain and mechanically forcing $\mathrm{PRG}\leq 0$ regardless of skill quality. Figure~\ref{fig:utility-ceiling}(a) shows that the ceiling rate varies substantially across configurations, ranging from $23.9\%$ under Claude Code\,/\,Sonnet~4 ($\mathrm{wo}=0.647$) to $50.3\%$ under Claude Code\,/\,Sonnet~4.6 ($\mathrm{wo}=0.799$), consistent with the pattern that stronger backbones saturate more scenarios. Figure~\ref{fig:utility-ceiling}(b) shows the PRG distribution across the six configurations. Three observations follow. \textbf{The PRG\,=\,0 bin dominates across all configurations}, and its mass is overwhelmingly attributable to ceiling scenarios on stronger backbones. \textbf{A non-trivial regression tail (PRG\,$<$\,0) is present in every configuration.} Regression rates range from $3.8\%$ (Claude Code\,/\,Sonnet~4.6) to $10.9\%$ (Claude Code\,/\,Sonnet~4), indicating that skills occasionally interfere with agent execution rather than merely failing to help. Notably, weaker backbones exhibit higher regression rates, suggesting that less capable agents are more susceptible to skill-induced interference. \textbf{Positive gains concentrate in the $(0,\,0.5]$ range} for most configurations, with a right tail in $(0.5,\,1]$ that is most pronounced under weaker baselines where ceiling effects are less dominant.

\subsection{PRG Stratified by Baseline Strength}
\label{app:utility-stratified}

\begin{figure}[!htbp]
  \centering
  \includegraphics[width=0.98\linewidth]{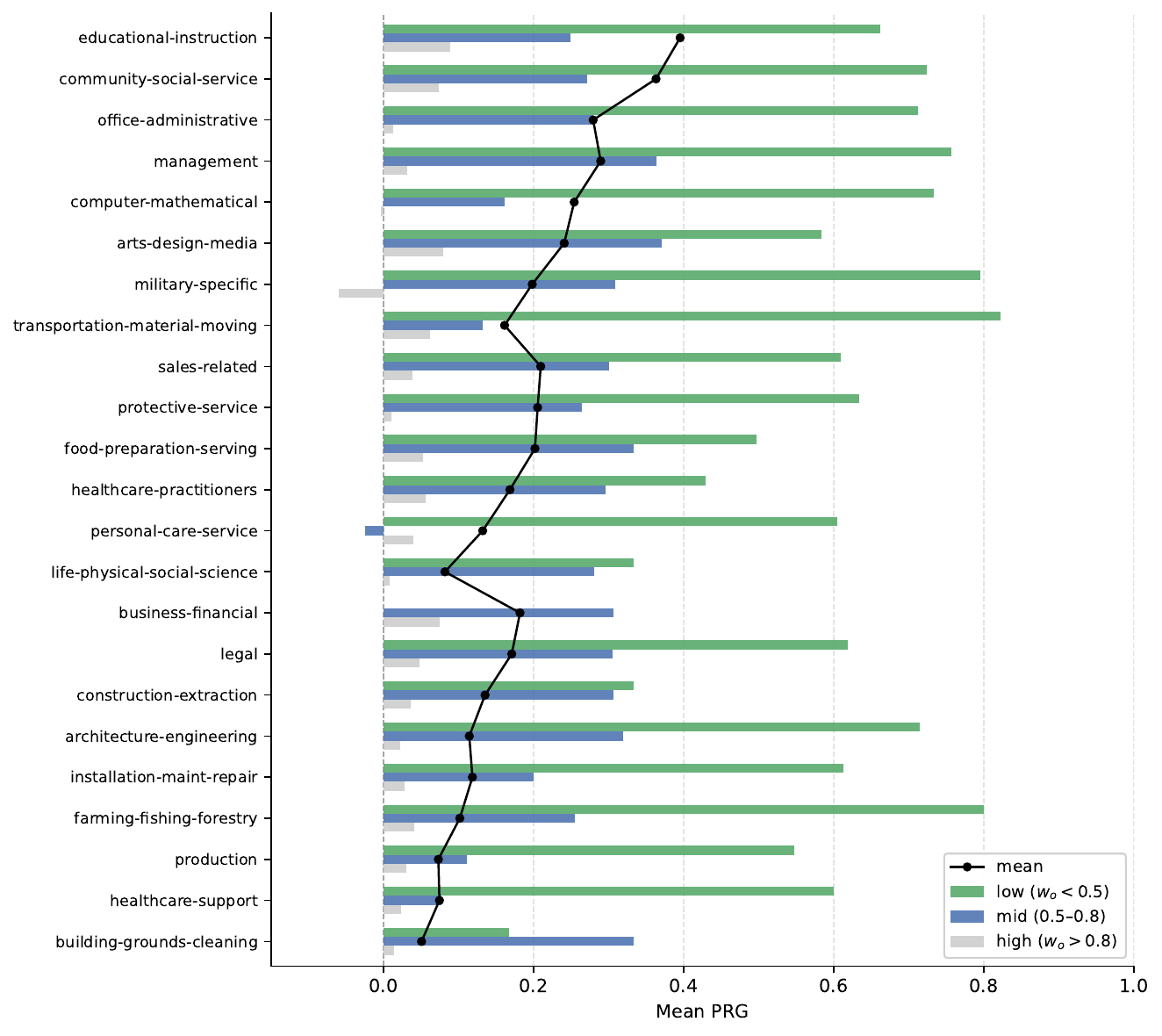}
  \caption{\textbf{Stratified mean $\mathrm{PRG}$ by occupational category.}
    Rows report the low, mid, and high baseline strata for
    Codex\,/\,GPT-5.4. The black line marks the unstratified category mean.}
  \label{fig:utility-stratified}
\end{figure}

We stratify Codex\,/\,GPT-5.4 scenarios by no-skill baseline pass rate into \emph{low} ($\mathrm{wo}<0.5$), \emph{mid} ($0.5\leq\mathrm{wo}\leq0.8$), and \emph{high} ($\mathrm{wo}>0.8$), as illustrated in Table~\ref{tab:stratum-summary}. The high stratum ($\mathrm{wo}>0.8$) accounts for $59.7\%$ of valid scenarios and dominates the unstratified mean ($\overline{\mathrm{PRG}}=0.183$). The low stratum ($\mathrm{wo}<0.5$, $n=93$) achieves a mean PRG of $0.648$, confirming that skills deliver substantial gains precisely where the backbone needs help the most. Figure~\ref{fig:utility-stratified} breaks down the same analysis by occupational category.
The low-stratum bars (green) are substantially longer than the
mid- and high-stratum bars in nearly every category, and the
high-stratum bars (grey) shrink to near zero across the board,
reflecting pervasive ceiling effects. The unstratified category mean (black line) is substantially lower than the low-stratum bars (green) across all categories, because the high stratum dominates the scenario count while delivering only modest gains. The stratum-conditional breakdown provides a more complete picture of skill contribution across different baseline regimes.

\begin{table}[!htbp]
  \caption{Stratum-conditional $\mathrm{PRG}$ summary for
    Codex\,/\,GPT-5.4 ($n=643$). \emph{zero\%} counts scenarios with $|\mathrm{PRG}|<0.001$,
    and \emph{neg\%} counts scenarios with $\mathrm{PRG}<-0.001$.}
  \label{tab:stratum-summary}
  \centering
  \small
  \begin{tabular}{lrrrrrr}
    \toprule
    \textbf{Stratum} & \textbf{$n$} & \textbf{rate} &
    \textbf{mean PRG} & \textbf{median PRG} &
    \textbf{zero\%} & \textbf{neg\%} \\
    \midrule
    Low ($\mathrm{wo} < 0.5$)             &  93 & 14.5\% & 0.648 & 0.667 &  2.2 & 0.0 \\
    Mid ($0.5 \leq \mathrm{wo} \leq 0.8$) & 166 & 25.8\% & 0.269 & 0.286 &  5.4 & 4.2 \\
    High ($\mathrm{wo} > 0.8$)            & 384 & 59.7\% & 0.034 & 0.000 & 67.2 & 4.7 \\
    \midrule
    All (unstratified)                    & 643 & 100\%  & 0.183 & 0.143 & 41.8 & 3.9 \\
    \bottomrule
  \end{tabular}
\end{table}

\FloatBarrier

\subsection{Category-wise PRG Distribution in One Configuration}
\label{app:utility-by-category}

\begin{figure}[!htbp]
  \centering
  \includegraphics[width=0.98\linewidth]{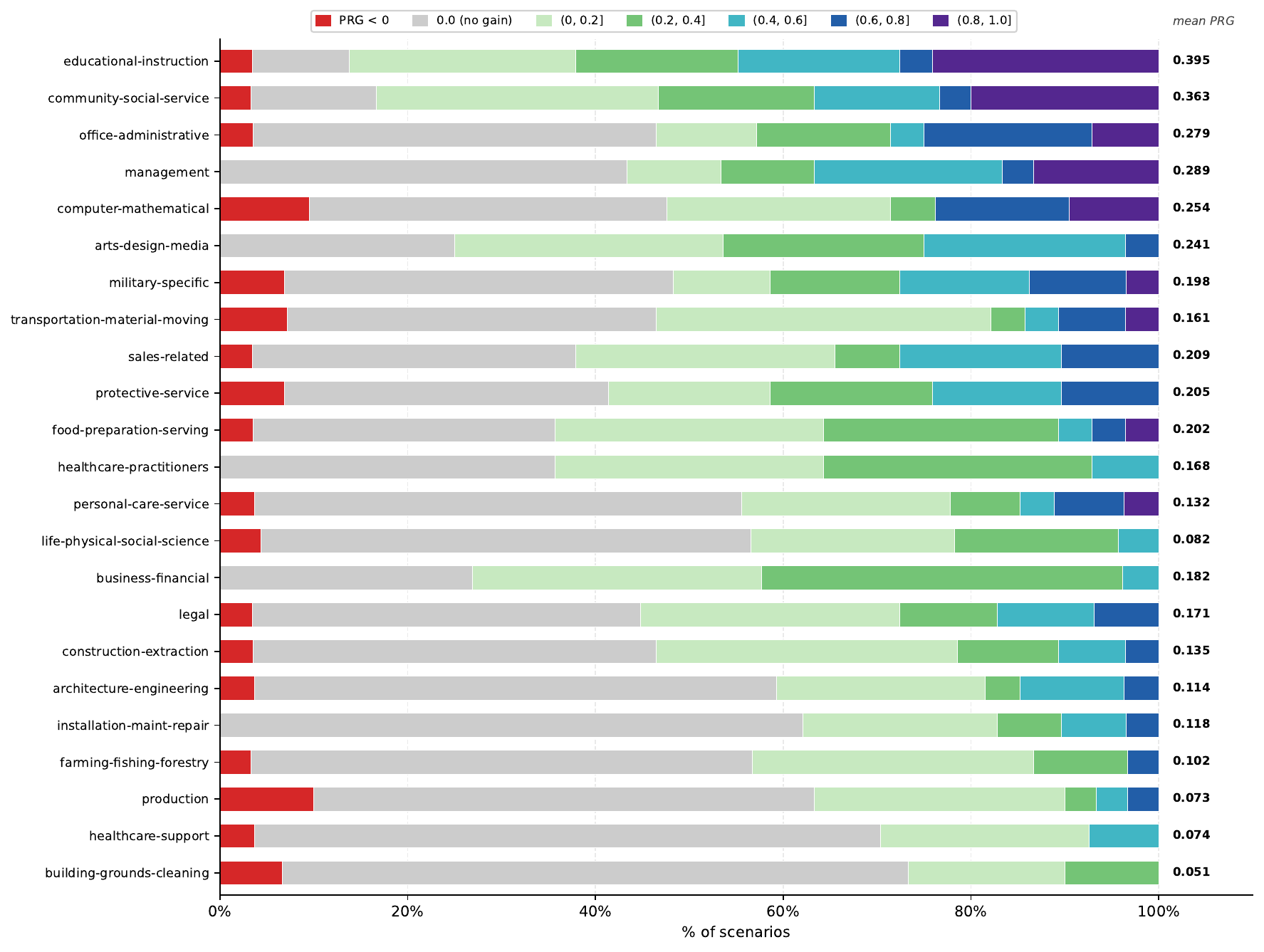}
  \caption{\textbf{Per-scenario $\mathrm{PRG}$ distribution by
    occupational category.} Bars are stacked by gain bucket.
    The leftmost red segment indicates scenarios where $\mathrm{PRG}<0$
    (skill-induced regression).
    Right-side labels report category mean $\mathrm{PRG}$.}
  \label{fig:utility-by-category}
\end{figure}

Figure~\ref{fig:utility-by-category} reports the full within-category distribution of scenario-level $\mathrm{PRG}$ for Codex\,/\,GPT-5.4, complementing the mean heatmap in Figure~\ref{fig:utility-cross-config}. The distribution separates three regimes that the category mean alone conceals: \emph{knowledge-structured} (high mean, small zero-gain share, e.g.\ \textit{educational-instruction} and \textit{community-social-service}), where skills consistently unlock new capabilities. \emph{bimodal} (moderate mean coexisting with a large zero-gain share and a right tail, e.g.\ \textit{office-administrative}, \textit{management}, and \textit{computer-mathematical}), where skills deliver large gains in some scenarios but are redundant in others. And \emph{operational} (low mean, majority zero-gain, e.g.\ \textit{healthcare-support} and \textit{building-grounds-cleaning}), where the baseline already handles most scenarios and skills add little on average.

\FloatBarrier



\newpage
\section{Safety Calibration}
\label{app:safety-calibration}

This appendix gives the calibration table used by the dynamic
security judge when converting behavioral evidence into
exploitability scores.

\begin{table}[!htbp]
  \caption{Exploitability calibration for dynamic safety findings.
    The security judge maps each dynamic verdict to the exploitability
    range used in Eq.~\ref{eq:safety-score}.}
  \label{tab:exploitability-calibration}
  \centering
  \small
  \begin{tabularx}{\linewidth}{p{4.2cm} X}
    \toprule
    \textbf{Verdict and score} & \textbf{Interpretation} \\
    \midrule
    \texttt{confirmed} (0.86--1.00) &
    End-to-end attack path is executed with observed evidence. \\
    \texttt{suspected} (0.51--0.85) &
    Strong behavioral signal exists, but evidence is incomplete. \\
    \texttt{agent\_refused} (0.50) &
    The agent blocks the risky path, but the package-level risk remains. \\
    \texttt{path\_exists\_not\_triggered} (0.20--0.49) &
    The path appears reachable but is not exercised in this run. \\
    \texttt{likely\_false\_positive} (0.10--0.19) &
    The trajectory shows little engagement with the predicted path. \\
    \texttt{no\_probe} (0.60) &
    No dynamic probe was executed. Exploitability defaults to a
    conservative prior. \\
    \bottomrule
  \end{tabularx}
\end{table}

\FloatBarrier

\newpage
\section{Safety Profile}
\label{app:safety-profile-details}

\begin{figure}[!htbp]
  \centering
  \includegraphics[width=0.92\linewidth]{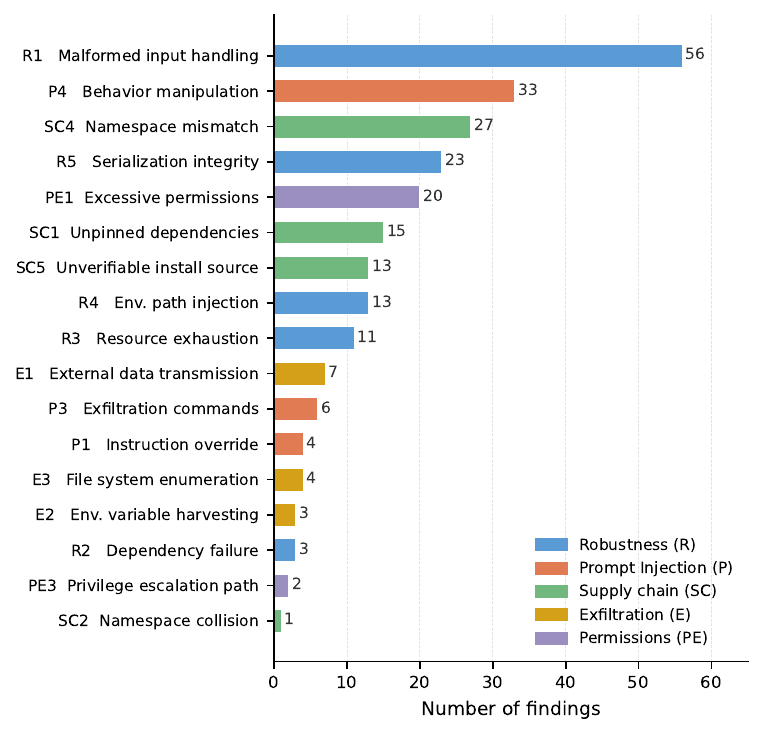}
  \caption{\textbf{Static-scan finding frequency by risk pattern.}
    Bars report the number of static findings for each risk pattern
    over the 226-skill set, color-coded by risk family: Robustness
    ($R$), Prompt Injection ($P$), Supply chain ($SC$), Exfiltration
    ($E$), and Permissions ($PE$).}
  \label{fig:risk-patterns}
\end{figure}

\begin{figure}[!htbp]
  \centering
  \includegraphics[width=0.78\linewidth]{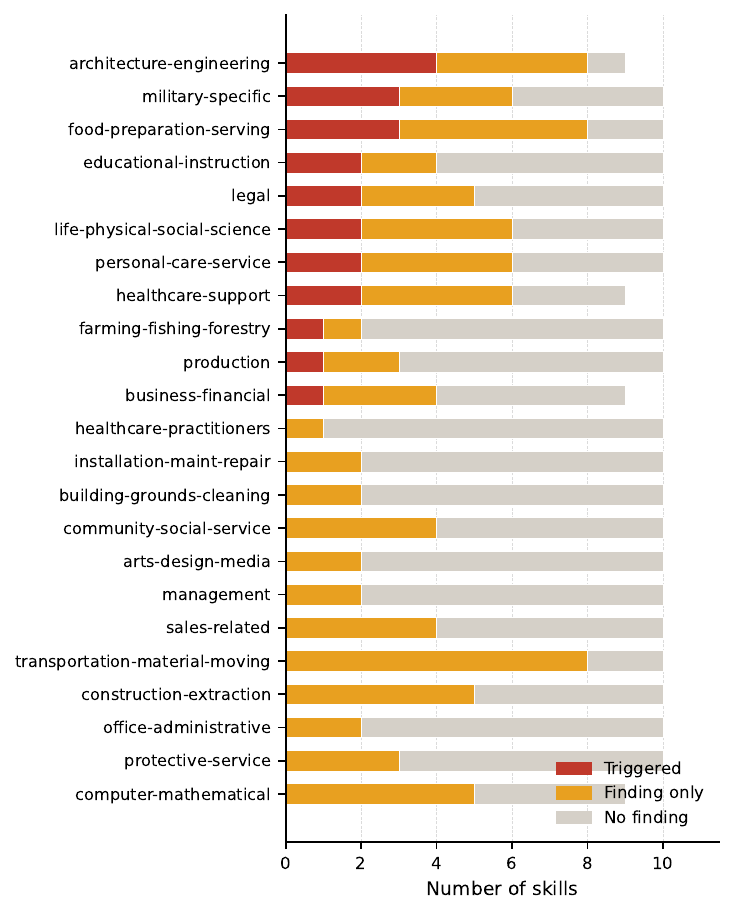}
  \caption{\textbf{Risk profile by occupational category.}
    For each of the 23 occupational categories, skills are
    partitioned into \emph{triggered} (a dynamic probe confirms
    exploitability), \emph{finding only} (the scanner reports at
    least one static finding but no dynamic probe triggers), and
    \emph{no finding}. Categories are sorted by the number of
    triggered skills.}
  \label{fig:occupation-profile}
\end{figure}

\begin{figure}[!htbp]
  \centering
  \includegraphics[width=0.78\linewidth]{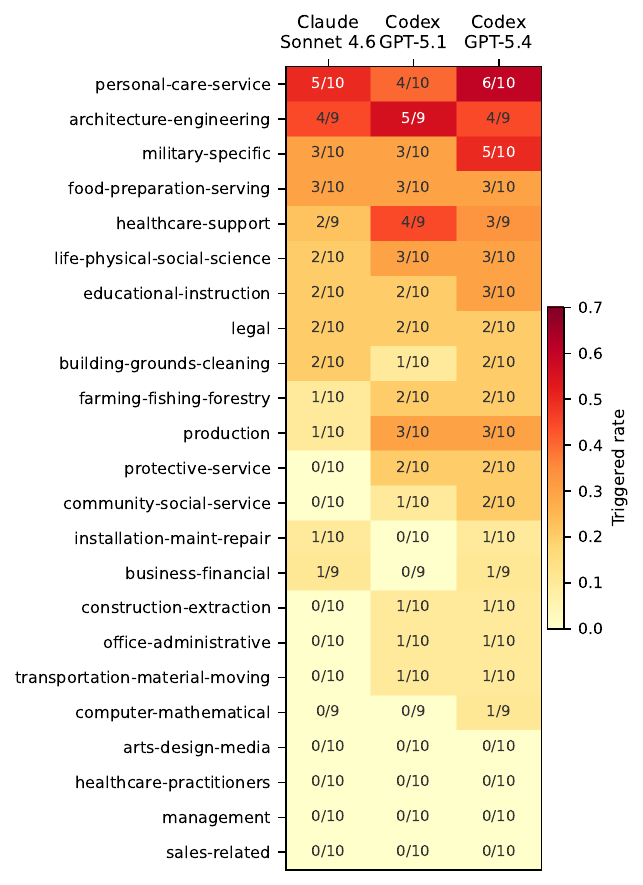}
  \caption{\textbf{Triggered rate by occupational category and
    agent.} Each cell reports triggered count over the number of
    tested skills in the category, on the same finding set as
    Figure~\ref{fig:safety-profile-main}(b). Rows are sorted by total
    triggered count across the three agents. Some categories remain
    consistently high-risk across agents, while others expose risk
    only under stronger execution backbones.}
  \label{fig:triggered-by-occupation-agent}
\end{figure}

This appendix complements the safety-profile analysis of Section~\ref{sec:exp-safety-profile}. The main text reports agent-conditioned dynamic outcomes in Figure~\ref{fig:safety-profile-main}; the figures below extend that analysis with pattern-level, occupation-level, and agent-level breakdowns. Figure~\ref{fig:risk-patterns} reports the static-finding frequency by risk pattern, and Figure~\ref{fig:occupation-profile} summarizes the aggregate occupational profile by separating triggered risks, finding-only risks, and skills with no finding. Figure~\ref{fig:triggered-by-occupation-agent} further breaks down the dynamic-probe triggered rate by occupational category and agent backbone, on the same fixed finding set used in Figure~\ref{fig:safety-profile-main}, so that cross-agent differences reflect agent behavior rather than differences in the underlying risk set.


\FloatBarrier
\section{Pattern-Level Scanner Diagnostics}
\label{app:per-pattern}
\begin{figure}[!htbp]
  \centering
  \includegraphics[width=0.92\linewidth]{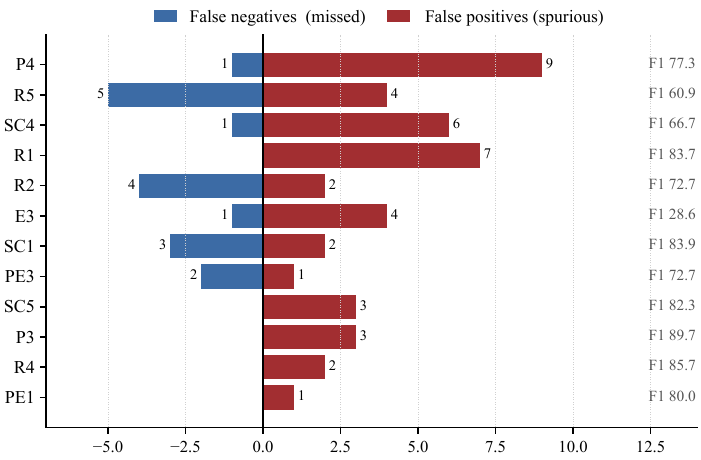}
  \caption{Per-pattern error decomposition for the 12 patterns with
    the largest total error ($\mathrm{FP}+\mathrm{FN}$). Negative
    bars indicate missed findings (false negatives); positive bars
    indicate spurious findings (false positives); pattern-level F1 is
    annotated on the right. The error profile is asymmetric: P4, R1,
    and SC4 are dominated by over-reporting, while R5, R2, and PE3
    are dominated by missed detections.}
  \label{fig:error-modes}
\end{figure}

Table~\ref{tab:per-pattern} reports the full per-pattern detection breakdown on the controlled subset of 29 skills. The pattern-level error decomposition in Figure~\ref{fig:error-modes} highlights the asymmetry between over-reporting and missed-detection error modes: P4, R1, and SC4 are dominated by false positives (over-reporting), while R5, R2, and PE3 are dominated by false negatives (missed detections).

\begin{table}[!htbp]
  \caption{Full per-pattern detection breakdown on the controlled
    subset of 29 skills.}
  \label{tab:per-pattern}
  \centering
  \footnotesize
  \begin{tabular}{llcrrrrrrrr}
    \toprule
    ID & Pattern & Sev & \#G & \#P & TP & FP & FN & P (\%) & R (\%) & F1 (\%) \\
    \midrule
    \multicolumn{11}{l}{\textit{Prompt Injection}} \\
    P1  & Instruction Override         & H & 11 & 11 & 11 & 0 & 0 & 100.0 & 100.0 & 100.0 \\
    P2  & Hidden Instructions          & H & 14 & 15 & 14 & 1 & 0 & 93.3  & 100.0 & 96.5  \\
    P3  & Exfiltration Commands        & H & 13 & 16 & 13 & 3 & 0 & 81.2  & 100.0 & 89.7  \\
    P4  & Behavior Manipulation        & M & 18 & 26 & 17 & 9 & 1 & 65.4  & 94.4  & 77.3  \\
    \midrule
    \multicolumn{11}{l}{\textit{Data Exfiltration}} \\
    E1  & External Data Transmission   & M & 16 & 17 & 16 & 1 & 0 & 94.1  & 100.0 & 97.0  \\
    E2  & Env Var Harvesting           & H & 8  & 9  & 8  & 1 & 0 & 88.9  & 100.0 & 94.1  \\
    E3  & FS Enumeration               & M & 2  & 5  & 1  & 4 & 1 & 20.0  & 50.0  & 28.6  \\
    E4  & Context Leakage              & H & 3  & 4  & 3  & 1 & 0 & 75.0  & 100.0 & 85.7  \\
    \midrule
    \multicolumn{11}{l}{\textit{Privilege Escalation}} \\
    PE1 & Excessive Permissions        & L & 2  & 3  & 2  & 1 & 0 & 66.7  & 100.0 & 80.0  \\
    PE2 & Sudo / Root Execution        & M & 1  & 1  & 1  & 0 & 0 & 100.0 & 100.0 & 100.0 \\
    PE3 & Credential Access            & H & 6  & 5  & 4  & 1 & 2 & 80.0  & 66.7  & 72.7  \\
    \midrule
    \multicolumn{11}{l}{\textit{Supply Chain}} \\
    SC1 & Unpinned Dependencies        & L & 16 & 15 & 13 & 2 & 3 & 86.7  & 81.2  & 83.9  \\
    SC2 & External Script Fetching     & H & 8  & 9  & 8  & 1 & 0 & 88.9  & 100.0 & 94.1  \\
    SC3 & Obfuscated Code              & H & 3  & 3  & 3  & 0 & 0 & 100.0 & 100.0 & 100.0 \\
    SC4 & Namespace Mismatch           & L & 8  & 13 & 7  & 6 & 1 & 53.8  & 87.5  & 66.7  \\
    SC5 & Unverifiable Install Source  & M & 7  & 10 & 7  & 3 & 0 & 70.0  & 100.0 & 82.3  \\
    \midrule
    \multicolumn{11}{l}{\textit{Robustness}} \\
    R1  & Malformed Input Handling     & L & 18 & 25 & 18 & 7 & 0 & 72.0  & 100.0 & 83.7  \\
    R2  & Network Timeout Missing      & L & 12 & 10 & 8  & 2 & 4 & 80.0  & 66.7  & 72.7  \\
    R3  & Retry / Loop Bounds          & L & 2  & 3  & 2  & 1 & 0 & 66.7  & 100.0 & 80.0  \\
    R4  & Error Handling               & M & 6  & 8  & 6  & 2 & 0 & 75.0  & 100.0 & 85.7  \\
    R5  & Resource Cleanup             & L & 12 & 11 & 7  & 4 & 5 & 63.6  & 58.3  & 60.9  \\
    \bottomrule
  \end{tabular}
\end{table}

\FloatBarrier
\clearpage

\end{document}